\documentclass[pdflatex]{sn-jnl}
\title{latexdiff}
\usepackage{cite}
\usepackage{xr}
\makeatletter
\usepackage{graphicx}
\usepackage{multirow}
\usepackage{amsmath,amssymb,amsfonts}
\usepackage{amsthm}
\usepackage{mathrsfs}
\usepackage[title]{appendix}
\usepackage{xcolor}
\usepackage{textcomp}
\usepackage{manyfoot}
\usepackage{booktabs}
\usepackage{algorithm}
\usepackage{algorithmicx}
\usepackage{algpseudocode}
\usepackage{listings}

\usepackage[section]{placeins}

\begin{document}

\title{Generating realistic global precipitation fields from modelled atmospheric circulation 

\thanks{Contact: Michael Aich, michael.aich@tum.de} 
}

\author*[1,2]{ \fnm{Michael} \sur{Aich}}\email{michael.aich@tum.de}
\author[1,2]{ \fnm{Sebastian} \sur{Bathiany}}
\author[1,2]{\fnm{Philipp} \sur{Hess}}
\author[1,2]{ \fnm{Yu} \sur{Huang}}
\author[1,2,3]{ \fnm{Niklas} \sur{Boers}}

\affil[1]{\small Technical University of Munich, Germany; Munich Climate Center; TUM School of Engineering and Design, Department of Aerospace and Geodesy, Earth System Modelling Group}
\affil[2]{\small Potsdam Institute for Climate Impact Research, Potsdam, Germany}
\affil[3]{\small Global Systems Institute and Department of Mathematics, University of Exeter, Exeter, UK}

\abstract{Improving the representation of precipitation in Earth system models (ESMs) is critical for assessing the impacts of climate change and especially of extreme events like floods and droughts. In existing ESMs, precipitation is not resolved explicitly, but represented by parameterizations. These typically rely on resolving approximated but computationally expensive column-based physics, not accounting for interactions between locations. They struggle to capture fine-scale precipitation processes and introduce significant biases. We present a novel approach, based on generative machine learning, which integrates a conditional diffusion model with a UNet architecture to generate accurate, high-resolution (0.25°) global daily precipitation fields from a small set of prognostic atmospheric variables. Unlike traditional parameterizations, our framework efficiently produces ensemble predictions, capturing uncertainties in precipitation, and does not require fine-tuning by hand. We train our model on the ERA5 reanalysis and present a method that allows us to apply it to unseen ESM data, enabling fast generation of probabilistic forecasts and climate scenarios. By leveraging interactions between global prognostic variables, our approach provides an alternative parameterization scheme that mitigates biases present in the ESM precipitation while maintaining consistency with its large-scale (annual) trends. This work demonstrates that complex precipitation patterns can be learned directly from large-scale atmospheric variables, offering a computationally efficient method to obtain high-resolution precipitation without the cost of running the dynamical model at such high resolution.}

\maketitle

\section{Introduction}\label{chapter1}

Heavy precipitation events will increase in frequency and magnitude in response to global warming \cite{IPCC_AR6_2023, prein2017future}. Extreme precipitation amplifies flood risks and leads to significant infrastructure, economic, and ecological damages \cite{davenport2021contribution}. Earth system models (ESMs) are used to simulate precipitation in current and future scenarios to help us understand changes and their impacts on the Earth system and society. Accurate high-resolution precipitation simulations are important for managing the risks of extreme precipitation events by guiding adaptation and risk mitigation measures, but still remain a major challenge in ESMs. 

However, simulating precipitation accurately at global scales is difficult because of the wide range of physical processes involved, from microphysics at the cloud scale to large-scale dynamics of the atmosphere. ESM model simulations are still extremely expensive and time consuming to run, as they need to numerically solve complex partial differential equations describing those processes. This greatly restricts the ability to create large ensemble projections for different climate change scenarios for the next century. 

ESMs represent precipitation as diagnostic variable that is computed from the dynamically resolved prognostic atmospheric variables. ESMs need to approximate the atmospheric circulation on discretized grids with often coarse spatial resolution. This leads to problems in correctly resolving small-scale dynamics that are important for inferring precipitation. Precipitation is a highly complex process and can not be directly resolved by the coarse ESM, but has to be parameterized. Parameterizations are simplified representations of the subgrid-scale processes that cannot be explicitly resolved due to computational constraints. At each time step, the ESM solves the atmospheric dynamics and applies parameterizations to account for subgrid-scale processes like convection and precipitation, updates the atmospheric state, and proceeds to the next step.

Traditional parameterization schemes are often column-based, treating each ESM grid column independently. While this approach is well-justified for many sub-grid processes \cite{arakawa2004cumulus}, it is a simplification that can struggle with processes that depend on horizontal context. For example, neglecting feedback between neighboring columns makes it difficult to capture mesoscale convective organization \cite{rio2019ongoing}. 
Additionally, the assumed timescales for convection closure can introduce further discrepancies, as convective processes in reality operate on a continuum of timescales that are not well represented in simple quasi-equilibrium closures\cite{rio2019ongoing}. The biases show in either over- or underestimation of precipitation intensity and especially in the ESMs' ability to represent extreme events. The bias is often particularly pronounced in the tropics, a prominent example being the double Intertropical Convergence Zone (ITCZ) bias \cite{hwang2013link}, where models produce excessive precipitation in the southern tropics.

We must distinguish between different small-scale processes. Cloud microphysics (i.e. drop nucleation or coalescence), which occur at the micrometer scale, will effectively never be directly resolved in global models. Convection, however, could theoretically be resolved with sufficiently fine grid spacing (i.e. in cloud-resolving models), but doing so for large ensembles at the global scale remains computationally out of reach for the near future due to the prohibitive computational cost associated with such high resolution. 

Furthermore, many parameterizations are formulated as heuristic equations that may miss key processes and may be to some extent structurally inadequate \cite{schneider2017earth}. Parameterizations rely on parameter calibration based on observed climatologies, which is to some extent subjective \cite{hourdin2023toward}. This limits their ability to generalize to changing climate conditions \cite{Bordoni2025} and can create inter-model disagreements \cite{schneider2017earth}.

The combination of coarse spatial discretization, simplifications like treating grid columns independently, inadequate functional forms in parameterizations, and subjective tuning all contribute to persistent biases in ESM-simulated precipitation. This makes impact assessments and water resource and flood management, which require precise spatial data at high resolution \cite{gutmann2014intercomparison}, extremely challenging. Also, ensembles of high-resolution precipitation projections are imperative for impact assessments but are currently not available due to computational limitations  \cite{Bordoni2025}. \newline

Recent studies have shown remarkable success utilizing Deep learning (DL) models for weather-forecasting \cite{pangu, gencast, lam2023learning, bodnar2024aurora, AIFS} and long-term climate simulations \cite{meyer2023ace}. Generative machine learning (ML) methods have been applied for downscaling, bias correction and forecasting tasks, adding the ability to produce ensembles that reflect intrinsic variability\cite{gencast,aich2024conditional, hess2024fast, addison2022machine, hess2022physically}. Most prominent are generative adversarial networks (GANs) \cite{goodfellow2020generative} and more recently diffusion models \cite{ho2020denoising}. GANs were among the first generative machine learning methods that we successfully applied to weather and climate data \cite{harris2022generative, leinonen2020stochastic}. Diffusion models (DMs) have in recent years often been preferred over GANs because their training is more stable and converges more easily. Also, GANs are prone to suffer from mode collapse, an issue where they are not able to approximate the full distribution, another area where DMs are superior.  \\

There have been efforts using deep learning to replace traditional column-based physical parameterizations in atmospheric models, while preserving the column structure \cite{rasp2018deep, gentine2018could, yuval2020stable}. In these approaches, more accurate but very expensive simulations (e.g. cloud-resolving or large-eddy simulations) serve as training data, capturing small-scale processes that standard parameterizations approximate. Neural networks are then trained to emulate complex subgrid-scale physics for much less computational cost. For instance, in ClimSim \cite{yu2023climsim} each atmospheric column can be treated as an independent regression task, mapping one-dimensional state variables to one-dimensional subgrid-scale tendencies. Despite promising results, a central limitation is that as these methods are trained on simulations they inherit biases present in them. As the neural network training is performed under specific climate model conditions, Deep learning  based parameterizations may struggle to generalize to changing climate states in decadal to centennial projections. As the simulations are very expensive, there is limited training data, which can hurt the generalization performance of data-hungry state-of-the-art DL methods. A recent approach \cite{kochkov2024neural} combined a fully differentiable dynamical core with a neural network to emulate small scale processes in a hybrid model setup designed for weather prediction. In a follow up work the authors introduce an additional neural network that infers precipitation rate from the atmospheric column state for a 2.8° grid spacing of satellite-based precipitation \cite{yuval2024neural}. 

We propose an alternative to column-based parameterizations, a probabilistic ML-based model that directly predicts high-resolution precipitation from low-resolution global atmospheric fields. A notable difference to column-based methods is that ML-based models can act at global scales to capture spatial interactions between various locations and also between different variables. Our model is trained exclusively on the ERA5 reanalysis, which integrates a wide range of observational data through advanced data assimilation techniques within a high-resolution (0.25°) weather model. This eliminates the need for emulating expensive small-scale simulations with nested cloud resolving models. Additionally, our proposed approach can also operate as a post-processing tool to derive refined precipitation fields from the atmospheric state once a ESM run has concluded. \\

We present a two-stage ML framework that operates on a small set of global 2-dimensional atmospheric variables, rather than on individual columns, to produce high-resolution daily precipitation at 0.25° (for a schematic overview see Fig. \ref{fig:main_fig}). Our goal is to show that such a framework can serve as a computationally efficient, probabilistic alternative to modeling precipitation that can reduce biases and capture realistic fine-scale spatial patterns. While our model is not intended as an operational replacement for fully coupled 3D parameterizations that predict tendencies, it demonstrates the potential for generative machine learning models to replace certain traditional parameterizations.

\section{Methods}\label{chapter2}

\begin{figure}[H]
    \centering
    \includegraphics[width=\textwidth]{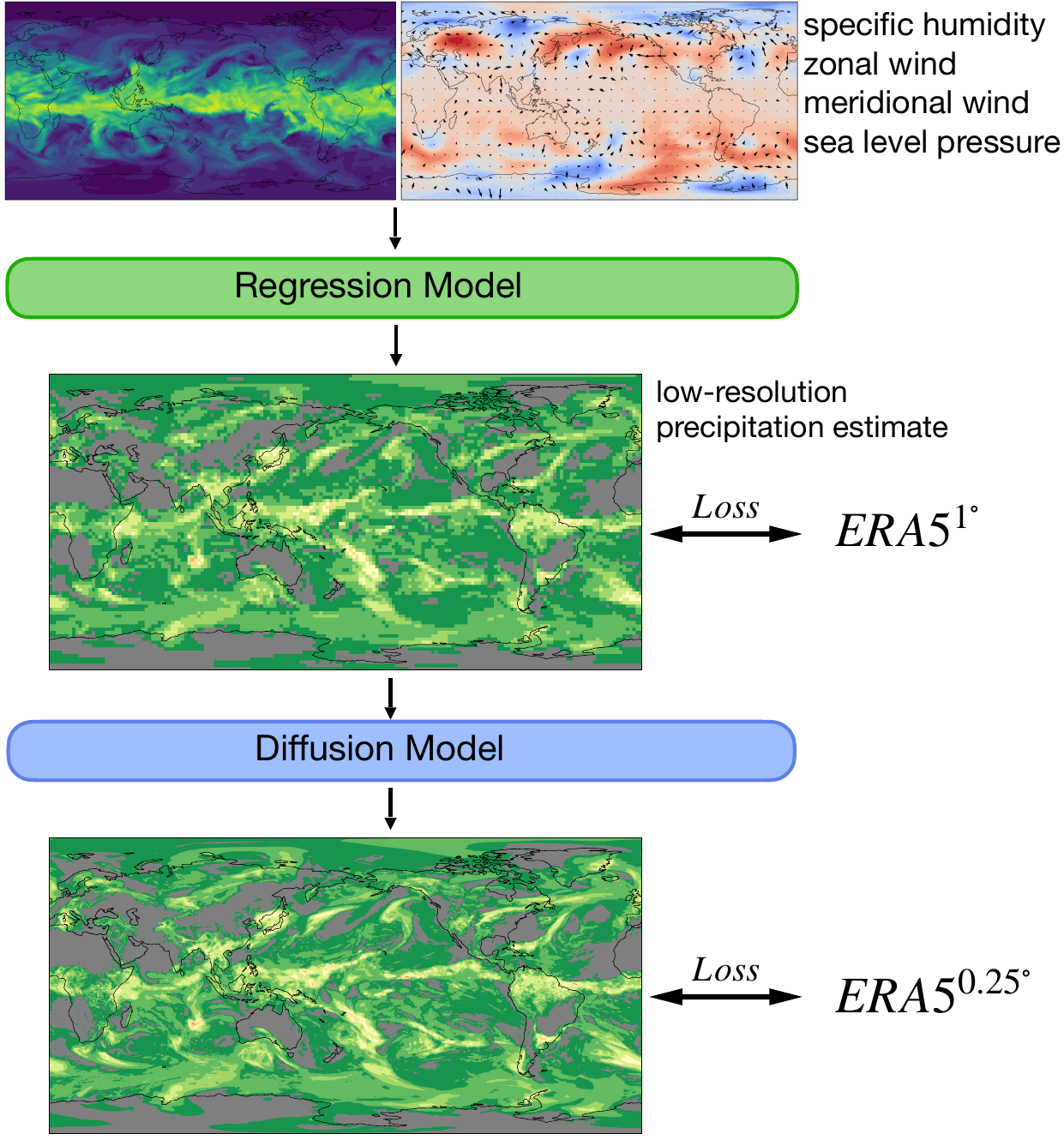}
    \caption{ \textbf{Schematic overview of our two‐stage approach for generating high‐resolution global precipitation from atmospheric variables.} 
    \textbf{1) Deterministic UNet Regression Model.} We first train a UNet model to learn the mapping from four atmospheric variables at 1° resolution (specific humidity at 850 hPa, near‐surface eastward and northward wind components at 10m, and sea‐level pressure) to 1° precipitation. During training, both the inputs and the precipitation targets come from ERA5 reanalysis. 
    \textbf{2) Generative diffusion model.} We train a conditional diffusion model to generate high-resolution (0.25°) precipitation fields, conditioned on coarse precipitation. To create this training condition, we interpolate ERA5 fields from 0.25° to 1° and inject noise. The diffusion model is trained to restore small‐scale spatial patterns of the high‐resolution ERA5 precipitation target. At inference we take atmospheric variables from the ESM and apply the trained UNet regression model to generate a 1° deterministic precipitation estimate. Before we use this estimate as a condition for our DM, we apply quantile delta mapping. The diffusion model is then conditioned on noisy quantile mapped precipitation estimates. }
    \phantomsection\label{fig:main_fig}
\end{figure}

Our framework consists of a two-stage deep learning pipeline: (1) a deterministic regression model to estimate 1° precipitation from 1° atmospheric variables and (2) a conditional diffusion model to perform probabilistic downscaling and bias correction from 1° to 0.25° (Fig. \ref{fig:main_fig}). Both models are trained exclusively on ERA5 reanalysis data. At inference we apply the trained models to ESM data, with specific domain adaptation steps (Quantile Delta Mapping and adding noise) to bridge the gap between the reanalysis and ESM data distributions.

\subsection{Datasets and Pre-processing}

\subsubsection*{Datasets}
\textbf{ERA5.} 
ERA5 \cite{hersbach_era5_2020} is a state-of-the-art atmospheric reanalysis dataset provided by the European Center for Medium-Range Weather Forecasting (ECMWF). We use ERA5 for training both of our models. The training period is 1980-01-01 to 2018-05-02, with 2018-05-03 to 2020-07-09 used for evaluation. \\

\textbf{GFDL-ESM4.}
For inference, we use data from the GFDL-ESM4 model \cite{dunne_gfdl_2020}, developed by the NOAA Geophysical Fluid Dynamics Laboratory (GFDL). GFDL-ESM4 is a member of Phase 6 of the Coupled Model Intercomparison Project (CMIP6). We use the historical period (1970-01-01 to 2014-12-31) and the SSP3-7.0 scenario \cite{oneill2016scenariomip} (2015-01-01 to 2100-12-31). For simplicity, we refer to the GFDL-ESM4 model output as GFDL from now on.

\textbf{MPI-ESM.} 
We also evaluate our framework on the MPI-ESM-HR model \cite{muller2018higher}. We use daily data at a spatial resolution of 0.9375°$\times$0.9375°, spanning from 1970-01-01 to 2014-12-31. For simplicity, we will from now on refer to the MPI-ESM-HR data as MPI. \\

\subsubsection*{Data pre-processing}
We use four prognostic variables on daily resolution as model inputs: specific humidity at 850 hPa, eastward (U) and northward (V) wind components at 10m, and mean sea level pressure (MSLP). The target variable is daily total precipitation in mm/day. We pre-process the data prior to the train/test split:

\begin{itemize}
\item GFDL data is bi-linearly interpolated from its native 1°$\times$1.25° grid to a 1°$\times$1° grid.
\item MPI data is bi-linearly interpolated from its native 0.9375°$\times$0.9375° grid to a 1°$\times$1° grid.
\item For precipitation we first add +1 mm/day to the data and then apply a log-transformation (following \cite{hess2024fast}).
\item All variables are projected to the range [-1, 1] to stabilize training.
\item High-resolution (0.25°) ERA5 data is bi-linearly interpolated to 1° (cdo command "remapbil") to create target data for training the regression model. 
\end{itemize}

\subsection{Stage 1: deterministic regression model}

\textbf{Objective:} The training objective is to learn a mapping from 1° daily atmospheric variables to 1° daily precipitation.

\textbf{Architecture:} We use the memory-efficient UNet architecture \cite{saharia2022photorealistic}. Specifically, the UNet has four input channels, one channel per atmospheric variable, and one output channel for precipitation. We increase the feature dimension across four downsampling stages using multipliers of (1,2,2,2), incorporating a ResNet \cite{he2016deep} block at each stage, and include only a single bottleneck attention layer to keep computational demand minimal. This progressively reduces the spatial resolution to capture larger-scale patterns, while simultaneously increasing the number of feature channels. The model has approximately 24 million parameters. Training and inference were performed on a single Nvidia H100 GPU with 80~GB of RAM.

\textbf{Training:} The model is trained on ERA5 data for 2000 epochs with the Adam optimizer \cite{kingma2014adam}, using a learning rate of $1e^{-4}$. Training for one epoch takes around 3 minutes, while inference of 1 field takes around 9 milliseconds. The 1° input data ($180\times360$ grid cells) is padded (zero-padding) to $192\times 368$ grid cells to be compatible with the UNet's downsampling blocks. After doing inference with our model, we index the precipitation output over the original $180\times360$ grid cells ERA5 domain. The padding and cropping is a purely technical choice that does not introduce artifacts at the prime meridian.

\subsection{Stage 2: probabilistic diffusion model} \label{train_dm}

\textbf{Objective:} The model learns to generate high-resolution (0.25°) precipitation fields, conditioned on a low-resolution (1°) precipitation field. 

\textbf{Architecture:} We use a Denoising Diffusion Probabilistic Model (DDPM) \cite{ho2020denoising} built upon the Efficient U-Net architecture \cite{saharia2022photorealistic}, downscaling from a $180 \times 360$ grid cell condition to a $720 \times 1440$ grid cells output. The diffusion model architecture utilizes a cosine schedule for noising the target data and a linear schedule for the condition during noise condition augmentation. The model has approximately 17 million parameters. Training and inference were performed on a single Nvidia H100 GPU with 80~GB of RAM.

\textbf{Training:} The model is trained for 1100 epochs with the Adam optimizer, a batch size of 1, a learning rate of $1e^{-4}$, and 1000 noising steps. Training for one epoch takes around 44 minutes, while inference of 1 field takes around 2 seconds. The diffusion model is trained only on ERA5 data. The 1° precipitation fields that serve as the condition are now bi-linearly interpolated to 0.25° to match the grid cell count ($720 \times 1440$ grid cells) of the 0.25° precipitation target. To match the data distribution expected during inference, we add a fixed level of Gaussian noise to the 1° ERA5 condition during training (Sec \ref{inference_DM}). We add Gaussian white noise that is independent for every grid cell following the forward process of the diffusion model. The added noise corrupts small-scale information in the condition. There is a direct relationship between the magnitude of the added noise and the spatial scales that are corrupted by the noise. Adding more noise (noise with larger amplitude) replaces the original information with noise up to larger spatial scales, which can be seen in a power spectral density (PSD) plot \cite{aich2024conditional,hess2024fast}. The DM learns to extract large-scale information that is unaffected by the noise and use this information to constrain its conditional generation task. The DM learns to generate the small-scale patterns (which we replaced with noise) in accordance to the large-scale spatial patterns that are unaffected by the noise. The model learns to extract large-scale information from the condition, which is helpful for generating the high-resolution target (as it shares the large-scale patterns with the condition).

\subsection{Inference pipeline and domain adaptation} 
\label{inference_DM}
At inference, the goal is to predict precipitation from the atmospheric variables of an ESM. This is a domain adaptation task, as ESM data is biased compared to ERA5 and the train and inference data distribution are not identically distributed. It is crucial to the ML model's generalization performance that training data and inference data are independent and identically distributed (i.i.d.). Our inference pipeline enables our trained ML model to predict precipitation using atmospheric variables from an ESM distribution, rather than the ERA5 data used during training.

\subsubsection{Stage 1 inference and large-scale bias correction}
First, the pre-processed 1° GFDL atmospheric variables are fed into the trained regression model, producing a deterministic 1° precipitation estimate. This precipitation estimate inherits systematic biases from the GFDL inputs. To correct these large-scale biases, we apply computationally efficient Quantile Delta Mapping (QDM) \cite{cannon2015bias} to the precipitation output. We fit QDM using the historical period (1980–2014) by mapping the UNet regression output (from historical GFDL atmospheric variables) to the 1° ERA5 precipitation. This learned correction is then applied to the UNet predictions for both the historical period and the future scenarios (2015-2100). The QDM-corrected 1° fields serve as the input for the next stage.

\subsubsection{Stage 2 inference and small-scale distribution matching}
The 1° QDM-corrected precipitation estimate is bi-linearly interpolated to $720 \times 1440$ grid cells to match the training setup. These fields, however, lack the realistic small-scale spatial variability of reanalysis data. Because of that, the distribution of the diffusion models' train and inference data do not match.

To resolve this, we apply the same amount of noise as used during training (Sec \ref{train_dm}) to the regression models' interpolated precipitation output. Adding noise with the same magnitude as in training effectively removes the smooth, biased small-scale patterns from the regression output and replaces them with noise. After applying quantile delta mapping, bilinear interpolation and adding the correct amount of noise to the regression model's precipitation estimate, the data distribution effectively matches the DM's training condition — 1° ERA5 precipitation bi-linearly interpolated with added noise (see \cite{aich2024conditional} for details). The amount of noise that is added is chosen before training the diffusion model and represents a hyperparameter. To select this magnitude, we analyze the power spectra intersection of the UNet regression output (bi-linearly interpolated to 0.25°) and the ERA5 fields (interpolated to 1° and then bi-linearly interpolated back to 0.25°). We look at the intersection for the bi-linearly interpolated 1° regression output and bi-linearly interpolated 1° ERA5 because the diffusion model internally interpolates the low-resolution conditioning input to the target resolution before generating high-frequency details. Even after quantile delta mapping, the regression model output and ERA5 fields will align only until a certain threshold scale in the spatial power spectrum, which we call $s$. We call the spatial scales that are above this point (around 900 km) large scales and the spatial scales below $s$ small scales. The choice of $s$ determines which spatial scales and therefore which size of precipitation events we want to preserve and which we want to regenerate according to the ERA5 distribution. Regenerating small scale patterns that were overlaid by noise is effectively a bias correction, as long as the spatial patterns are preserved. Without preservation of some patterns in the input it would just be unconditional ERA5 emulation. For a detailed description of the choice of $s$ and how it effectively solves this domain adaptation problem we refer to \cite{aich2024conditional}.

Finally, after quantile delta mapping, bilinear interpolation and noising, the regression model's precipitation is fed into the trained diffusion model as a condition, which iteratively denoises completely noisy (Gaussian) field. The diffusion model regenerates ERA5-like, high-resolution (0.25°) small-scale patterns consistent with the bias-corrected large-scale structures of the condition. Our inference pipeline takes four atmospheric ESM inputs as condition to generate precipitation fields at a higher spatial resolution while simultaneously mitigating ESM biases. At inference, we use 100 noising steps (down from 1000 in training) which massively reduces inference time.

\section{Results}\label{chapter3}

To generate our results, we apply the two-stage inference framework detailed in Section \ref{chapter2}. First, the UNet regression model estimates 1° precipitation from GFDL atmospheric variables. We then apply Quantile Delta Mapping (QDM) to these fields to correct systematic large-scale biases. Unless otherwise stated, results labeled as 'UNet' or 'regression model' always refer to this QDM-corrected regression model output to allow for a fair comparison between UNet and DM outputs. The conditional diffusion model (DM) then uses this estimate to generate bias-corrected, 0.25° high-resolution ensembles. Visual inspection of representative snapshots (Fig. \ref*{fig:individual_samples}) confirms that the regression model matches the large-scale structure of the ERA5 validation ground truth well (Fig. \ref*{fig:individual_samples}A \& Fig. \ref*{fig:individual_samples}B). The individual samples show that the DM preserves the large-scale structures (Fig. \ref*{fig:individual_samples}C \& Fig. \ref*{fig:individual_samples}D and Fig. \ref*{fig:individual_samples}F \& Fig. \ref*{fig:individual_samples}G) while generating plausible high-resolution details, confirming that the DM output is correctly conditioned on the atmospheric state.

\subsection{Correction of spatial biases}

\begin{figure}[!htb]
    \centering
    \includegraphics[width=0.6 \textwidth]{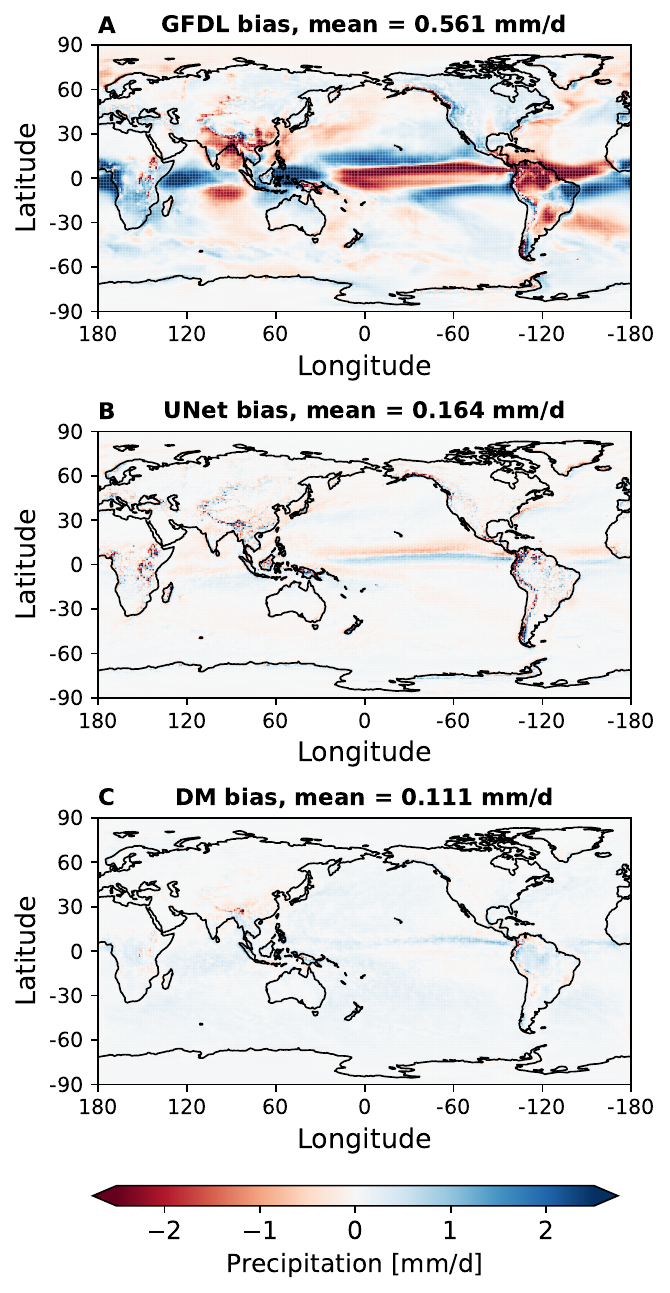}
    \caption{ \textbf{Comparison of model biases}. The maps show the precipitation biases of \textbf{(A)} GFDL, \textbf{(B)} the QDM UNet regression model, and \textbf{(C)} the diffusion model relative to ERA5. The difference map of GFDL shows pronounced bias in the tropical regions, including a double ITCZ. Our regression model alone already improves over GFDL, but the DM leads to further reduced deviations from ERA5 and the smallest mean absolute bias. Note that we interpolated ERA5 and the DM to 1° by applying average pooling for fair comparison to GFDL.}
    \phantomsection\label{fig:climatology_diff}
\end{figure}

The difference plots over 40 years (Fig. \ref{fig:climatology_diff}) reveal that GFDL (Fig. \ref{fig:climatology_diff}A) has by far the largest spatial bias, especially showcasing a strong double ITCZ bias, when compared to our regression model (Fig. \ref{fig:climatology_diff}B) and our DM (Fig. \ref{fig:climatology_diff}C). We aggregated both our DM as well as ERA5 from 0.25° to 1° with average pooling for better comparability. The absolute climatologies (Fig. \ref{fig:climatology}) further illustrate these discrepancies. The relative differences are shown in Figure \ref*{fig:relative_bias_gfdl}. GFDL has the highest mean absolute bias of 0.561 mm/day. The UNet regression model, even after applying QDM, yields a mean absolute bias of 0.164 mm/day. The diffusion model further reduces this bias to 0.111 mm/day. The GFDL precipitation is heavily biased in the tropics, with substantial regional over- and underestimation compared to ERA5. We also compare the climatologies of ERA5 (Fig. \ref*{fig:climatology_hr}A) and our DM precipitation (Fig. \ref*{fig:climatology_hr}B) at 0.25° high resolution (GFDL is only available at 1°). The climatologies over 40 years are very similar and only show small differences (Fig. \ref*{fig:climatology_hr}C). This is in line with the small mean absolute bias of 0.136 mm/day. Our alternative precipitation estimates and in particular the DM exhibit substantially lower spatial biases compared to GFDL at 1°. Our DM has also only a very small spatial bias at the finer 0.25° ERA5-resolution, where GFDL is not available.

\subsection{Improving precipitation statistics}

\begin{figure}[!htb]
    \centering
    \includegraphics[width=1.2\textwidth]{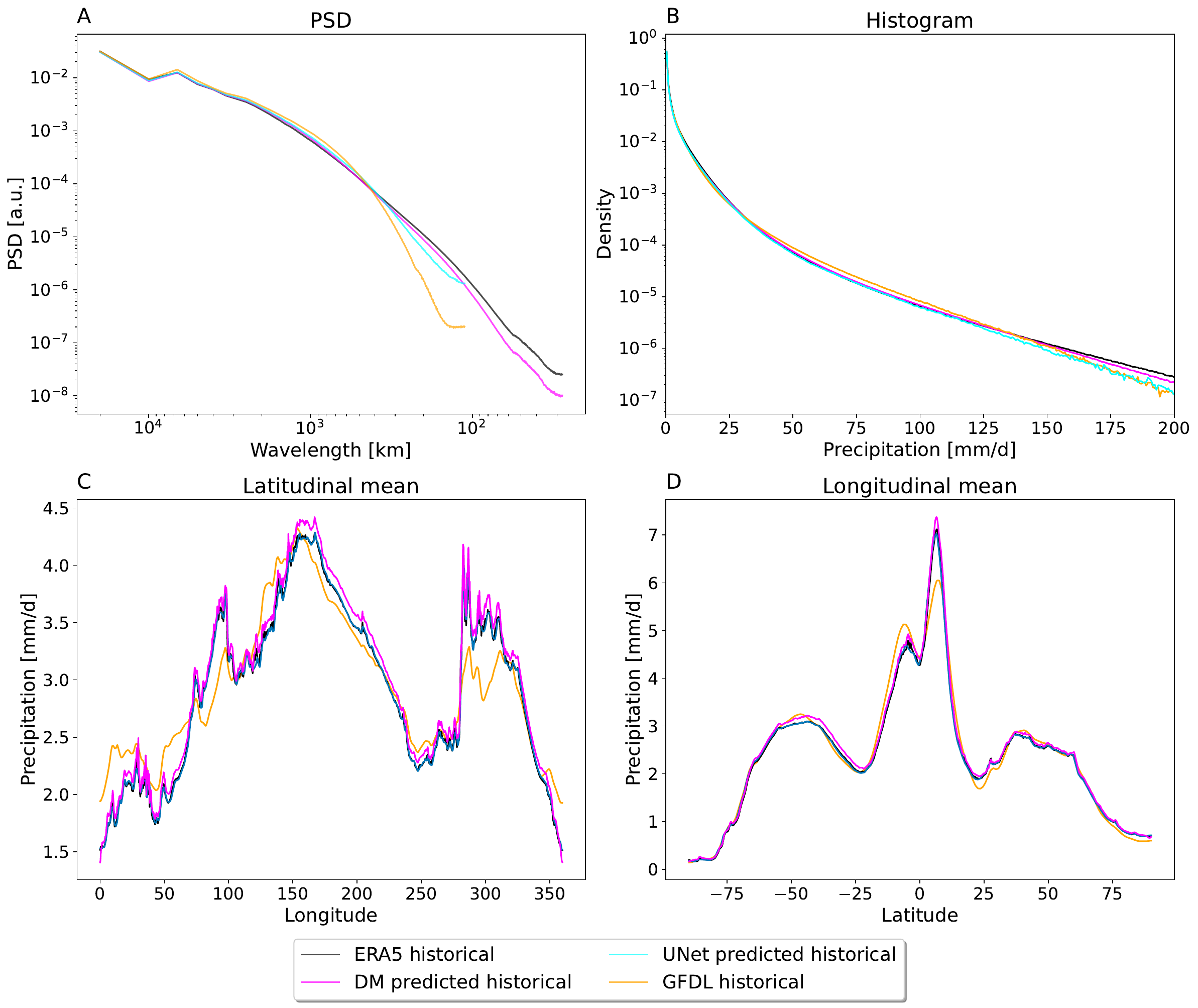}
    \caption{ \textbf{Reproduction of the statistics of ERA5 at 0.25° resolution over 40 years for the GFDL-ESM.} 
    \textbf{(A)} Mean spatial power spectral density (PSD).
    \textbf{(B)} Histogram indicating the precipitation frequencies.
    \textbf{(C)} Longitude profile, given by the data averaged over all latitudes, weighted by the cosine of latitude to account for the varying grid cell area.
    \textbf{(D)} Latitude profile, given by the data averaged over all longitudes.   
    For panels (B-C), we bi-linearly interpolate the GFDL and our regression model predictions of precipitation (orange/cyan) from 1° to the 0.25° to compare them against the 0.25° ERA5 and DM fields.
    }
    \phantomsection\label{fig:gfdl_eval}
\end{figure}

The spatial power spectral density (PSD) shows that GFDL's amplitude of variability on small spatial scales is substantially reduced compared to ERA5 (Fig. \ref{fig:gfdl_eval}A). The ESM produces blurry outputs lacking spatial variability at scales up to 400 km. At scales between 100 km and 400 km, our regression model is aligned better with ERA5. Our DM can generate precipitation fields at 0.25° resolution that align well with the spatial spectrum of ERA5, showing its capabilities in correctly modeling spatial variability. 

The histogram (Fig. \ref{fig:gfdl_eval}B) shows that the 1° GFDL model overestimates precipitation between 25 mm/day and 100 mm/day and underestimates precipitation greater than 150 mm/day compared to 0.25° ERA5. Our 1° UNet regression model represents the precipitation more accurately, but also slightly underestimates the small but more frequent precipitation events with less than 25 mm/day. Our 0.25° DM performs best with only a very slight underestimation of extreme events with more than 150 mm/day, similar to the regression model. We confirm the statistical robustness using a bootstrap analysis (Fig. \ref*{fig:bootstrapped_hist}). We performed the bootstrap analysis by resampling the 40 years of daily data with replacement 1,000 times. For each resampled dataset, we computed the precipitation histogram. The resulting aggregate histograms were used to calculate the 2.5th, 50th, and 97.5th percentiles for each bin. The 95$\%$ confidence intervals are generally narrow due to the large sample size, becoming visible only at the extreme tails ($>$170 mm/day) where event frequencies are low. Notably, the GFDL uncertainty envelope is wider than that of the DM or ERA5, because the coarser 1° resolution of GFDL yields fewer spatial data points compared to the 0.25° high-resolution references. 

For the latitudinal and longitudinal mean comparisons (Fig. \ref{fig:gfdl_eval}C and Fig. \ref{fig:gfdl_eval}D), we bi-linearly interpolated the regression model and GFDL precipitation to 0.25°. Both plots show that both of our models align well with the ERA5 statistics, while GFDL sometimes strongly deviates, depending on the region. Because the regression model outputs are post-processed with QDM, they match the ERA5 means almost perfectly. In contrast, the DM shows slight deviations as a result of its stochastic generative process, which prioritizes realistic spatial variability over perfect mean preservation. The differences between the DM, UNet and GFDL compared to ERA5 are further highlighted in Fig. \ref*{fig:gfdl_eval_diff}.

Furthermore, we demonstrate that our DM framework clearly outperforms a traditional statistical bias-correction and downscaling baseline (Fig. \ref*{fig:bilin_bench}), especially in recovering high-frequency spatial details. We further substantiated this by applying our pre-trained model to the MPI-ESM for inference without retraining. The results (Fig. \ref*{fig:mpi_benchmark}) confirm that our approach is not limited to a specific ESM, recovering missing small-scale details and improving the distribution of intermediate precipitation intensities compared to the raw MPI-ESM output (instead of GFDL-ESM4).

\subsection{Reproducing extreme events}

\begin{figure}[!htb]
    \centering
    \includegraphics[width=\textwidth]{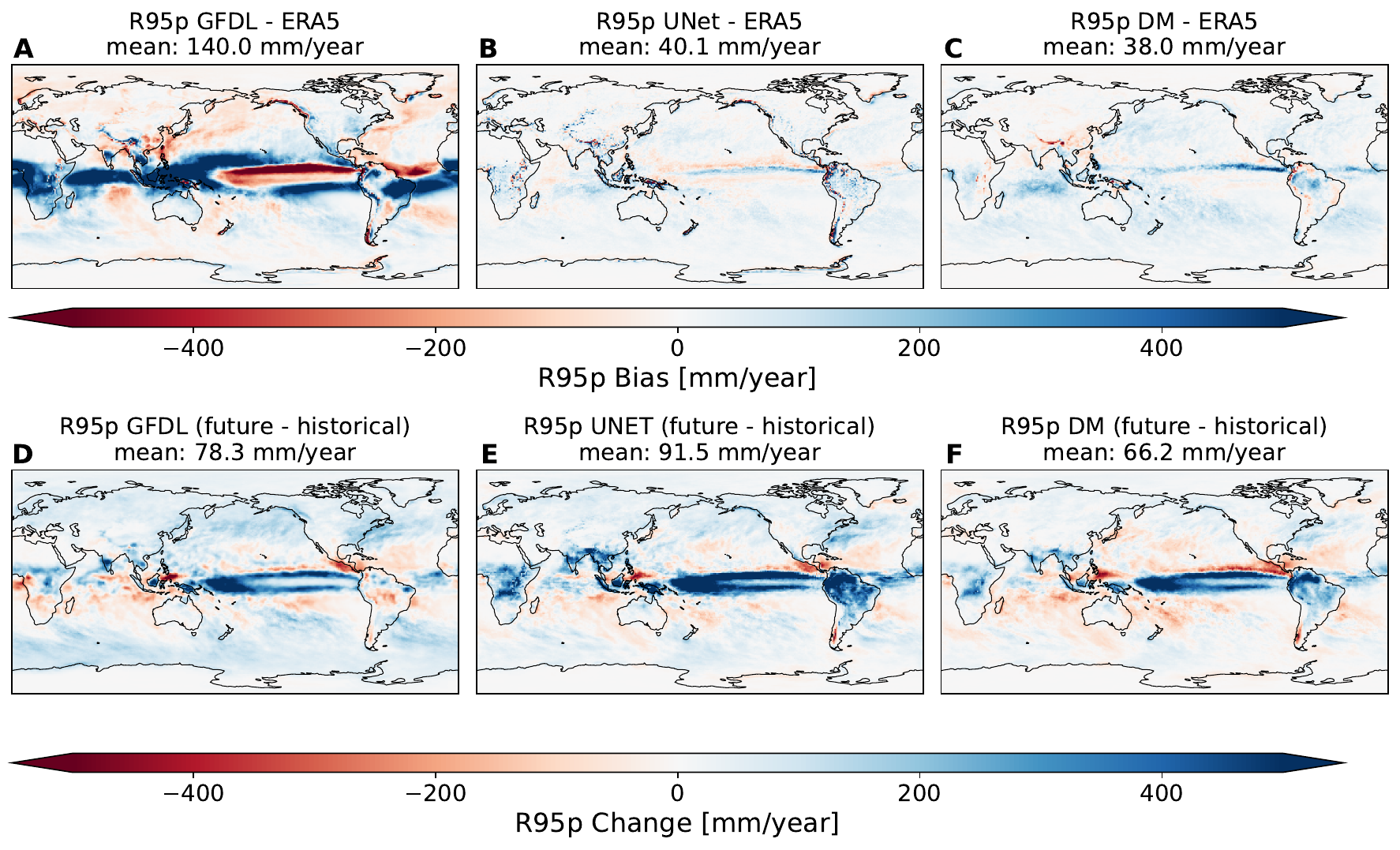}
    \caption{\textbf{Evaluating extreme event coverage for historical and future scenarios}. We compare the total precipitation from days exceeding the 95th percentile of daily rainfall (R95p). \textbf{(A-C)}: Bias in historical data (1980 - 2005). (A) GFDL bias (GFDL - ERA5), (B) UNet bias (UNet - ERA5 ), and (C) DM bias (DM - ERA5). ERA5 and the DM were interpolated to 1° to match the resolution of GFDL. Positive values (blue) indicate overestimation of extreme precipitation by the respective model, while negative values (red) indicate underestimation. The GFDL model exhibits strong wet biases in the tropics during the historical period (A), whereas the DM and UNet models show reduced biases (B, C).
    Bottom row \textbf{(D-F)}: Projected changes in R95p between the late 21st century (2075 – 2100, SSP3‐7.0) and the historical period. (D) GFDL future change (future – historical), (E) UNet future change, and (F) DM future change. Positive values (blue) indicate an intensification of extreme precipitation, while negative values (red) indicate reductions. Under future warming, all models predict an increase in R95p over the tropics. While the UNet predicts the largest mean increase, the DM effectively moderates this signal, resulting in a more conservative intensification compared to GFDL. The area-weighted mean absolute error (A–C) and mean change (D–F) are indicated in the subplot titles.
    }
    \phantomsection\label{fig:r95p_dif}
\end{figure}

To assess whether our diffusion model (DM) can adequately reproduce present-day climate extremes and adapt to unobserved future conditions, we compare the R95p extreme precipitation index across historical and projected climates under SSP3‐7.0 (Fig.~\ref{fig:r95p_dif}). The R95p index is computed as the annual total precipitation from days exceeding the 95th percentile. We use this index to compare historical extreme precipitation of ERA5, our UNet regression model, GFDL and our DM (Fig. \ref*{fig:r95p_clima}A - Fig. \ref*{fig:r95p_clima}D).

During the historical period (1980–2005), GFDL exhibits pronounced biases in the mid-latitudes and tropics, including a double ITCZ bias, leading to an overestimation of extreme rainfall around the equator (Fig.~\ref{fig:r95p_dif}A). Our ML models follow ERA5 closely, exhibiting smaller deviations in both spatial extent and magnitude. The DM in particular reduces the wet bias of GFDL effectively, and is more closely aligned with ERA5. The reference period for the computation is 1980–2020. To verify the robustness of these findings for rarer extremes, we also evaluated the R99.9p index (Fig. \ref*{fig:r99_difference}).
The results confirm that the DM maintains its skill for higher percentiles, also effectively correcting the tropical wet biases of the input while capturing the signal of future intensification.

We further evaluate extreme-event performance by comparing consecutive wet days (CWD) and consecutive dry days (CDD) during the historical period (1980-2005) against ERA5 (Fig.\ref*{fig:cdd_cwd_dif}). The largest differences tend to appear over the tropics, where pronounced convective precipitation leads to larger deviations in CWD, and over subtropical regions where prolonged dry periods cause sizable shifts in CDD. Both our ML-based approaches, but especially the diffusion model, capture ERA5 patterns of CWD and CDD more accurately, clearly improving upon the biases of GFDL.

\begin{figure}[!htb]
    \centering
    \includegraphics[width=\textwidth]{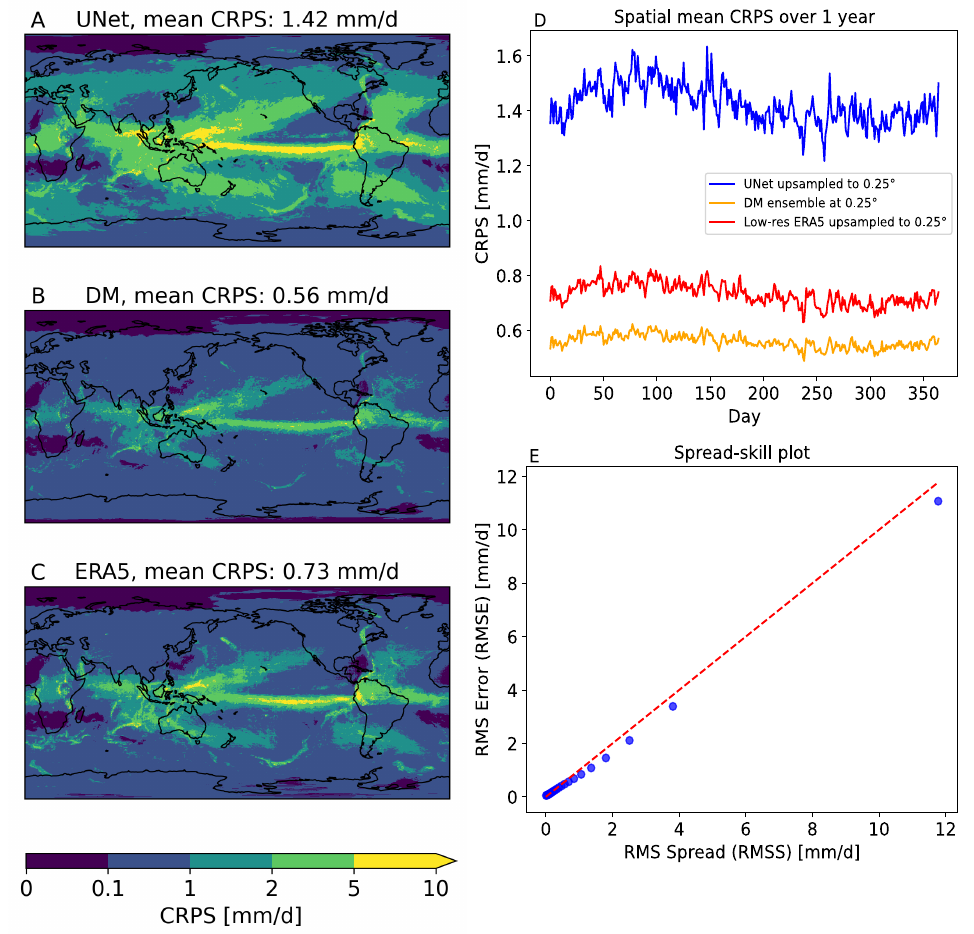}
    \caption{ \textbf{Evaluation of the 50 DM ensemble members over one reference ERA5 year.} Temporally averaged continuous ranked probability score (CRPS) (lower is better) for \textbf{(A)} applying our deterministic UNet, followed by QDM and then bi-linearly interpolating from 1° to 0.25°, \textbf{(B)} our DM and \textbf{(C)} the bi-linearly interpolated deterministic ERA5 baseline (from 1° to 0.25°). \textbf{(D)} Spatially averaged CRPS. \textbf{(E)} Spread-skill plot. Following the 1:1 line indicates well-calibrated spread of the ensemble.
    }
    \phantomsection\label{fig:crps}
\end{figure}

\subsection{Ensemble evaluation}

A key strength of our generative DM approach compared to traditional column-based parameterization is its ability to generate a diverse ensemble of possible, spatially consistent high-resolution (0.25°) precipitation realizations for a given low-resolution (1°) condition. This accounts for the aleatoric uncertainty inherent in inferring fine-scale details from coarse variables.

It is critical to evaluate the internal variability of our diffusion model in the downscaling task. We investigate whether the uncertainty in a DM ensemble aligns with the inherent uncertainty of downscaling the data from 1° to 0.25°. Note that the diffusion model also performs bias correction by regenerating noisy small-scale variability. 
We condition our DM on a 1° bi-linearly interpolated noisy ERA5 validation year and evaluate the DM 50 times for each day in the year, resulting in a 50-member DM ensemble of one-year trajectories. The downscaling ground truth will be the respective high-resolution ERA5 year. 
We compare the continuous ranked probability score (CRPS) \cite{crps} of our 50-member DM downscaling with 1° ERA5 and the QDM-corrected 1° regression models output both bi-linearly interpolated to 0.25°. For this specific experiment, the QDM correction was fitted using the UNet output derived from coarse (1°) ERA5 inputs to ensure a consistent evaluation.

This evaluation is conducted over a one-year period from the ERA5 validation dataset, starting on 2018-05-03. This comparison assesses how well the DM’s ensemble spread matches the inherent uncertainty. 
Our DM outperforms both deterministic baselines in terms of temporally averaged CRPS (Fig. \ref*{fig:crps}A - Fig. \ref*{fig:crps}C). Note that for deterministic baselines, the CRPS reduces to the mean absolute error. The mean CRPS of our DM is the lowest at 0.56 mm/day, better than that of the ERA5 interpolated from 1° to 0.25° (0.73 mm/day) and the interpolated QDM-corrected prediction (1.42 mm/day). The baselines effectively filter out the spatial variability between 0.25° and 1°, while our DM is able to reconstruct it. The spatially averaged CRPS (Fig. \ref*{fig:crps}D) also shows that our model has the best scores overall. The spread-skill relationship (calculated following \cite{haynes2023creating}) roughly follows the 1:1 line, indicating an accurate, well-calibrated representation of uncertainty (Fig. \ref*{fig:crps}E). For larger spreads the RMSE is slightly too low (underconfident). To verify that the diffusion model accurately captures large-scale precipitation patterns, we also evaluated performance at the 1° scale by interpolating the predictions (Fig. \ref*{fig:crps_1degre}). The relative performance ranking remains consistent with the high-resolution results. This confirms that the DM accurately captures large-scale structures while generating realistic small-scale variability.

\subsection{Precipitation generation for climate projections}

In the following, we apply our ML method to future climate projections. This is a challenging task, given that the underlying distribution shifts over time in response to changing climate conditions. A comparison between our regression model (bi-linearly interpolated to 0.25°) and the diffusion model (Fig. \ref*{fig:ssp_statisitcs}) for SSP3-7.0 data between 2015-2100 shows that our DM is still able to increase the spatial resolution of the precipitation predictions from 1° to 0.25° with ERA5-like large-scale patterns and small-scale deviations consistent with those observed in the historical period (Fig. \ref*{fig:ssp_statisitcs}A).  \\

\textbf{Trend preservation} \\

\begin{figure}[!htb]
    \centering
    \includegraphics[width=\textwidth]{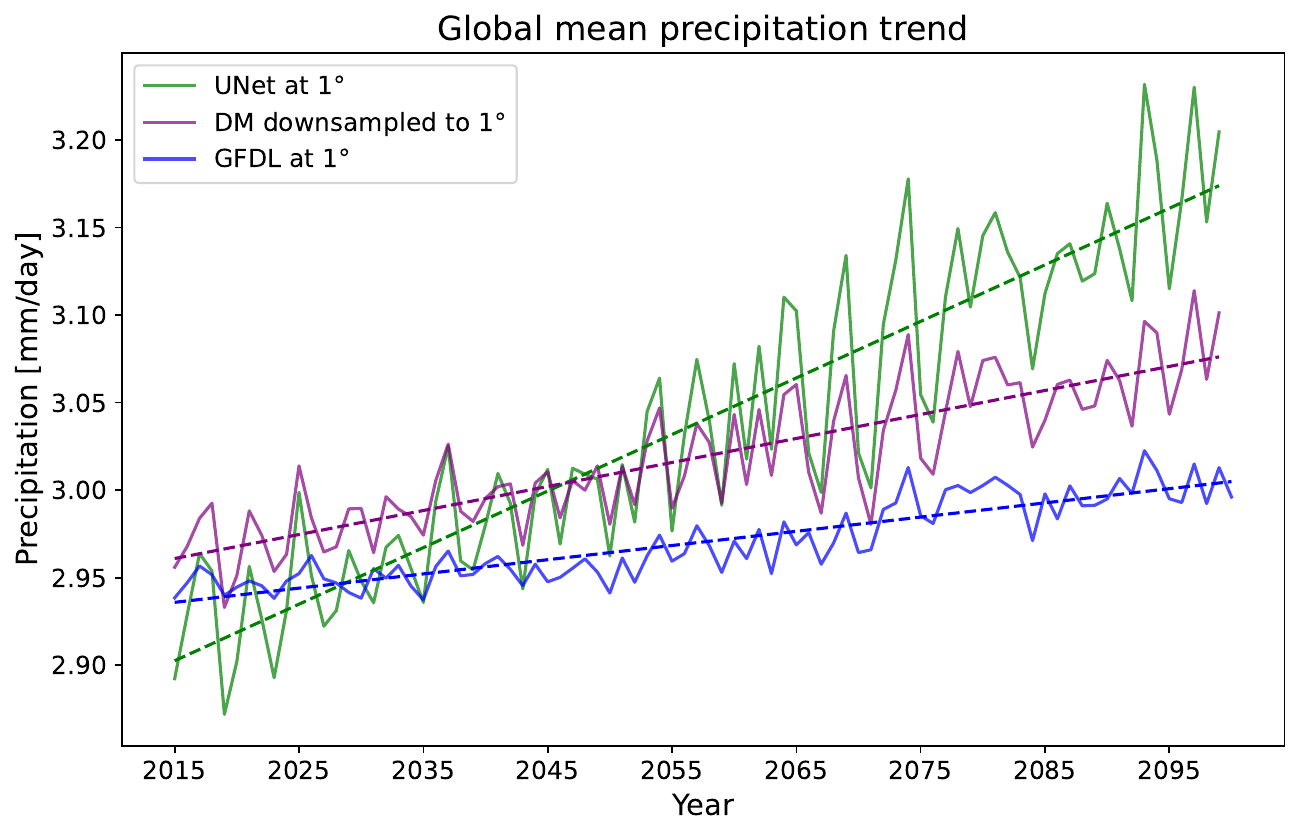}
    \caption{ \textbf{Precipitation trends under the SSP3-7.0 scenario}. Comparison of globally and annually averaged precipitation trends for the SSP3-7.0 scenario for GFDL (blue), our UNet regression model (green) and our DM (purple). For comparability, the DM was interpolated to 1°. The dashed lines represent the linear regression trends for each model. The spatial and globally averaged trend of the UNet (even after quantile delta mapping) differs from GFDL. The DM corrects the global average trend of the UNet, aligning it more with GFDL. The DM trend is slightly larger than that of GFDL, while the mean values are slightly lower. Note that the GFDL trend is not a strict ground truth but rather serves as a reference for comparison, as the DM performs a correction of the data towards the ERA5 distribution. 
    }
    \phantomsection\label{fig:trend_ssp370}
\end{figure}

We specifically tested the ability of our diffusion model to extrapolate to an unseen future climate scenario, namely SSP3-7.0, which projects relatively high greenhouse gas emissions and therefore increasing specific humidity. First, we initialize our regression model with the SSP3-7.0 atmospheric variables and then apply quantile delta mapping. We find that the regression model predicts an increasing precipitation trend, perhaps driven largely by the higher specific humidity (Fig. \ref{fig:trend_ssp370}). The spatial trend at each location (Fig. \ref*{fig:spatial_trend_ssp370}A - Fig. \ref*{fig:spatial_trend_ssp370}C) reveals that applying the DM on the QDM UNet precipitation preserves the spatial patterns of the trend. While the spatial patterns are similar, the DM corrects the local structures and intensities. The spatial patterns of trends predicted by the regression model differ in sign from the GFDL projections in some regions. The spatially averaged annual mean trend reveals that the UNet regression model has a different annual mean precipitation and exhibits an overly large variance, resulting in a too large trend. This high inter-annual variance is the direct result of a systematic spatial error at the daily fields. The regression model has to make predictions in an out-of-distribution scenario for future scenario GFDL fields and therefore produces fields with flawed spatial structures and unrealistic intensities. Quantile Delta Mapping cannot fix this problem entirely. While QDM can correct the long-term climatological average, by adjusting the mean precipitation, it cannot correct the flawed small-scale spatial physics. The DM however will generate correct small scale information, reducing the variance in the quantile delta mapped regression output and thereby correcting the overly steep annual mean trend. This is especially relevant for the far future (post 2055) where to UNet predictions have increasingly larger variability, resulting in a steep trend. When we correct the regression model's precipitation with our DM, the DM dampens the variance such that the annual precipitation trend matches the trend of the GFDL SSP3-7.0 projection more closely (Fig. \ref{fig:trend_ssp370}). After applying the DM, there is only a small deviation in the globally averaged mean values and trend compared to the GFDL projection. It is important to note that the GFDL simulation represents just one possible realization of a future climate scenario and acts as a reference rather than a ground truth target. Our method relies on the assumption that the statistical relationship between large-scale circulation and small-scale precipitation variability, as learned from ERA5, remains valid under future warming. Consequently, our DM robustly adapts to these future scenarios by preserving the large-scale structures projected by GFDL while generating consistent, ERA5-like small-scale variability. 

To verify that trend preservation is a robust feature of our framework and not limited to a specific model or time period, we analyzed the combined historical and future period from 1970–2100 for both GFDL-ESM and the independent model MPI-ESM-HR (Fig. \ref{fig:full_trend}). The evaluation confirms that trend preservation is maintained across the full 1970–2100 period for both ESMs. For both ESMs, the quantile delta mapped regression models produces unrealistically high variance (and thus steeper trends) which the DM consistently corrects. Crucially, the experiment reveals that our framework does not simply reproduce the ERA5 trend during the historical period. Instead, the DM-corrected trend aligns with the respective ESMs rather than forcing the data to match the steeper historical trend of ERA5, confirming that our framework adapts to the specific input climate signal rather than overfitting to the training distribution. 
Consequently, our framework allows us to correct spatial patterns while preserving the overarching global trend information of the driving atmospheric variables. We note, however, that the output of our generation routine does not strictly replicate the native precipitation trend of the driving ESM. The regression model effectively infers precipitation by mapping the climate change signal of the ESMs atmospheric variables through the observational relationships learned from ERA5. The generative diffusion step is crucial because the intermediate quantile delta mapping applied to the initial regression predictions introduces an unrealistically high inter-annual variability (see Fig. \ref{fig:full_trend}). This variance inflation is an artifact that occurs when grid-wise quantile mapping is applied to overly smooth deterministic model outputs \cite{maraun2013bias}. The diffusion process naturally dampens this QDM artifact. The native ESM projections in Fig. \ref{fig:trend_ssp370} and Fig. \ref{fig:full_trend} are therefore shown for reference rather than as an absolute ground truth. Accordingly, our final DM-corrected precipitation is not expected to perfectly match the original ESM trend. Overall, Fig. S\ref{fig:full_trend} demonstrates that our framework yields consistent predictions for two different ESMs with the slopes of the DM-corrected lines remaining largely consistent with the respective reference projections.

We note that although SSP3-7.0 is an out-of-distribution test in terms of temperature increases, in practice it represents a moderate distribution shift for precipitation, characterized primarily by an increased frequency of high values rather than entirely novel conditions outside the historical range. \\

\textbf{Spatial climate change signal} \\

Fig. \ref*{fig:rel_dif_past_future} shows the relative signal of climate change in precipitation (2075-2100 vs. 1980-2005) relative to the historical period for GFDL (Fig. \ref*{fig:rel_dif_past_future}A), the UNet regression model (Fig. \ref*{fig:rel_dif_past_future}B), and our diffusion model (Fig. \ref*{fig:rel_dif_past_future}C). Overall, both UNet and DM capture large-scale changes while offering more moderate local responses than GFDL. In the tropics and especially the ITCZ region, GFDL exhibits a stronger positive precipitation response compared to both data‐driven methods, which generally temper the climate change signal. In particular, in parts of South America, the sign of the projected change switches from positive in GFDL to almost zero or slightly negative for the UNet and DM. As we show that the overall trend of precipitation is well approximated in statistical metrics and time series trends, this could suggest that the ML-based models correct for some of the potential biases of GFDL, resulting in different and, in some cases, opposite regional precipitation trends.

Under the future warming scenario SSP3‐7.0 (between 2075–2100), all models predict an increase in R95p in the tropics relative to the historical period (Fig. \ref{fig:r95p_dif}D – Fig. \ref{fig:r95p_dif}F). Because model biases can alter the magnitude of these changes, we compare the sign of the change across the plots, which is largely consistent among GFDL, UNet regression, and DM. Overall, the UNet projects the largest mean intensification of precipitation extremes. However, the DM produces a more moderate increase than both UNet and GFDL, with some regional disagreements, particularly over South America. The double ITCZ bias remains visible in the future projections of GFDL. Visually, both GFDL and DM show stronger extremes within the tropical belt, although GFDL changes are larger in magnitude overall, possibly reflecting its known wet bias. Our DM reproduces similar spatial patterns (signs of change) but with a more modest amplitude, suggesting that it partially corrects the climatological biases of GFDL, while retaining the overarching signal of climate change.

\section{Discussion and Conclusions}\label{chapter4}

We introduced a two-stage framework for inferring high-resolution global precipitation fields from a small set of atmospheric variables, composed of a deterministic UNet regression model together with a probabilistic conditional diffusion model. Our generative Machine Learning approach hence provides a spatial-pattern-aware alternative to column-based parameterization of precipitation for ESMs. 

We chose this two-stage framework to address the distributional mismatch between the relatively smooth large-scale atmospheric inputs and the highly intermittent, fine-scale precipitation target. We found that learning this complex mapping in a single generative step is very challenging. This difficulty was also identified by \cite{mardani2025residual}, who employ a two-stage approach to address this.
A key advantage of our approach is that both our models are trained exclusively on a single dataset, in our case ERA5 reanalysis data. This training strategy ensures that the framework is not tied to the specific biases or parameterizations of any specific ESM. The only parameter in the training of the ML-models that depends on the ESM is the amount of noise that is added to the DM's condition. While applying the framework to ESMs with significantly different resolution or biases requires retraining with a noise level calibrated to that specific ESM, we empirically confirmed that for ESMs with similar spectral characteristics (GFDL-ESM4 and MPI-ESM), our model works well without retraining. Not changing the noise-scale could potentially result in incomplete removal of small-scale biases in models where the spatial scales are resolved worse. \\

Our setup aligns with the so-called perfect model approach \cite{van2023deep}, i.e. learning to reconstruct high-resolution target fields from artificially coarsened input fields. Recent studies \cite{van2023deep, rampal2024enhancing} suggest that the imperfect model approach, mapping low-resolution ESM output directly to high-resolution observations, can be advantageous for implicitly correcting ESM biases. However, it requires temporal correspondence between input and target data. Such paired data is available when using regional climate models (RCMs) that are driven by ESM boundary conditions. Since our problem setup involves mapping a free-running ESM to ERA5 reanalysis, which represent two independent realizations of internal variability without alignment, the imperfect strategy is not applicable. \\

Another advantage of our framework is its computational efficiency. With inference times of roughly two seconds per global 0.25° precipitation prediction (on a single Nvidia H100 GPU), our framework is fast compared to running an ESM at 0.25° resolution. 

This efficiency is partly due to our deliberate choice of a minimal set of 2D predictor variables. While traditional parameterizations rely on the full vertical profiles, our results suggest that modern deep learning models can extract implicit information about the 3D atmospheric state from surface fields. 
For instance, MSLP captures the integrated mass of the atmospheric column, while our surface winds are intrinsically coupled to the larger-scale 3D circulation. Our evaluation confirms that these 2D fields contain sufficient information to reconstruct high-resolution precipitation patterns without explicitly providing the vertical dimension.

Our approach efficiently produces ensembles of high-resolution precipitation fields from low-resolution (1°) atmospheric variables, which faithfully represent the inherent uncertainty in the downscaling process. Our method can also be used in post-processing for generating large ensembles and long-term precipitation projections, which are essential for impact and risk assessments. \\

Our evaluation demonstrates that our method substantially reduces the bias compared to our ESM reference. Our approach successfully avoids the prominent double ITCZ bias which is strongly pronounced in the GFDL model. Our method also reproduces accurate small-scale spatial patterns with which ESMs often struggle. In the power spectral density we confirm that the small-scale spatial variability of the downscaled precipitation aligns closely with the ERA5 reanalysis data taken as ground truth, which is a big improvement over the comparably blurry GFDL precipitation fields. The spatial variability of small scales is essential for accurate simulation of regional precipitation extremes and associated impact assessments.  \\

While ERA5 is used as the ground truth target in this study, we acknowledge that reanalysis precipitation data contains inherent biases. Specifically, ERA5 has a wet bias and can underestimate the intensity of extreme precipitation events \cite{lavers2022evaluation}. However, given its global coverage, high resolution and physical consistency, ERA5 remains the standard dataset for training machine learning models. Our framework is data-agnostic and can be retrained on observational datasets for specific operational applications. \\

Importantly, our framework preserves the climate change signal inherent in the specific humidity under the SSP3-7.0 scenario, while correcting small-scale spatial patterns in accordance with reanalysis data. Compared to the quantile delta mapped UNet regression model, our diffusion model adjusts the spatial variability of precipitation more effectively, bringing it closer in line with the GFDL projections, while the global mean remains slightly different. This deviation reflects the fact that GFDL should not be considered as ground truth in our setup, and the diffusion model performs a bias correction toward ERA5, balancing consistency with climate model trends and alignment with observational data. In contrast, our UNet regression model alone fails to reproduce the mean and predicts too high variance. An essential part of our method is to add noise to the regression model’s 1° output before conditioning the diffusion model on it. This removes the flawed small-scale features while preserving large-scale dynamics, effectively acting as a bias correction for the small-scales. We show that the diffusion model preserves the spatial patterns of the trend (i.e. regions of increase and decrease) predicted by the UNet, while refining the magnitude of these changes by regenerating realistic small-scale variability. By regenerating improved, reanalysis-like, small-scale structures, the DM aligns the global, spatially averaged, trend with the trend of the GFDL projection, reducing the precipitation intensities of the UNet predictions. Our model demonstrates robustness to out-of-distribution scenarios, as shown by its ability to preserve trends and to capture extreme events under the SSP3-7.0 scenario. Overall, evaluating the model's performance on future scenarios is challenging due to the lack of high-resolution ground truth. Future research could compare against dynamically downscaled datasets like Regional Climate Models (RCMs), although such models introduce their own biases. 

Generally, we acknowledge that our model generates reanalysis-like fine-scale details based on the historical training data. This implicitly assumes that the relationship between large-scale states and small-scale structures remains largely constant under climate change. Investigating this relationship could be an avenue of research in future work. \\

Our method is trained on a frame-by-frame basis, having no information from neighboring time steps. The temporal consistency of the large-scale precipitation patterns is inherited directly from the evolution of the atmospheric variables, which give us temporal continuity for large-scale features. However, we acknowledge that the fine-scale features generated by the diffusion model are stochastic and not explicitly constrained to be temporally consistent from one time step to the next. It is worth noting, however, that such consistency is, in the case of precipitation, primarily relevant for high-frequency data (e.g. sub-daily) where substantial autocorrelation exists. For our daily precipitation fields, the temporal autocorrelation of these fine-scale features is less pronounced compared to sub-daily scales. While our approach effectively captures spatial statistics and large-scale dynamics, extending the framework to a video diffusion model to enforce temporal continuity for fine-scale structures remains a valuable direction for future development, for applications involving sub-daily data. \\

An important practical benefit of our framework producing 0.25° resolution is that it enables coupling the precipitation outputs to impact or land models without requiring a higher resolution for the 1° atmospheric variables. This decoupling of scales can substantially lower computational costs while still providing the detailed precipitation fields needed for impact studies. \\

Several avenues for future research remain. An  extension is to integrate our DM-based parameterization directly into a climate model and evaluate its stability in a fully coupled system. Such experiments can help determine whether further constraints on the UNet outputs are necessary when deployed in operational settings. This includes incorporating physical constraints to assure that conservation laws are fulfilled \cite{harder2023hard}, or adopting alternative loss functions to better capture specific features such as precipitation extremes \cite{doury2024suitability}.\\

Bridging the gap from our proof of concept to a fully operational parameterization would also require extending the framework to predict 3D tendencies, potentially incorporating vertical profiles and ensuring its coupling with other physical processes. This is non-trivial given the known challenges regarding stability and drift for this task \cite{brenowitz2018prognostic, yuval2020stable}. Optimizing the inference strategy with techniques such as model distillation \cite{luhman_knowledge_2021} could be explored to accelerate the sampling process and further reduce computational cost. 
\clearpage

\bibliography{sn-bibliography}

\newpage

\section*{Acknowledgments}

\subsection*{Funding}
Funded under the Excellence Strategy of the Federal Government and the L\"ander through the TUM Innovation Network EarthCare. \\
This is ClimTip contribution \#140; the ClimTip project has received funding from the European Union's Horizon Europe research and innovation programme under grant agreement No. 101137601: Funded by the European Union. Views and opinions expressed are however those of the author(s) only and do not necessarily reflect those of the European Union or the European Climate, Infrastructure and Environment Executive Agency (CINEA). Neither the European Union nor the granting authority can be held responsible for them.\\
The Past to Future (P2F) project has received funding from the European Union’s Horizon Europe research and innovation programme under grant agreement No. 101184070: Funded by the European Union. Views and opinions expressed are however those of the author(s) only and do not necessarily reflect those of the European Union or the European Climate, Infrastructure and Environment Executive Agency (CINEA). Neither the European Union nor the granting authority can be held responsible for them.\\
PH, SB, and NB acknowledge funding by the Volkswagen Foundation.\\
YH acknowledges funding by the Alexander von Humboldt Foundation.\\

\subsection*{Author contributions}

\subsection*{Data Availability}

The ERA5 reanalysis was accessed from \url{https://developers.google.com/earth-engine/datasets/catalog/ECMWF_ERA5_DAILY} \\
The GFDL-ESM4 and MPI-ESM-HR data is available at \url{https://esgf-data.dkrz.de/search/cmip6-dkrz/}. \\

\subsection*{Code Availability}
The code will be made available on GitHub at \url{https://github.com/aim56009/precipitation_from_atmosphere.git} by the time of publication.

\subsection*{Competing interests}
The authors declare no competing interests.

\clearpage

\newpage
\renewcommand{\thefigure}{S\arabic{figure}} 

\section*{Fig. S1}

\begin{figure}[!htb]
    \centering
    \includegraphics[width=\textwidth]{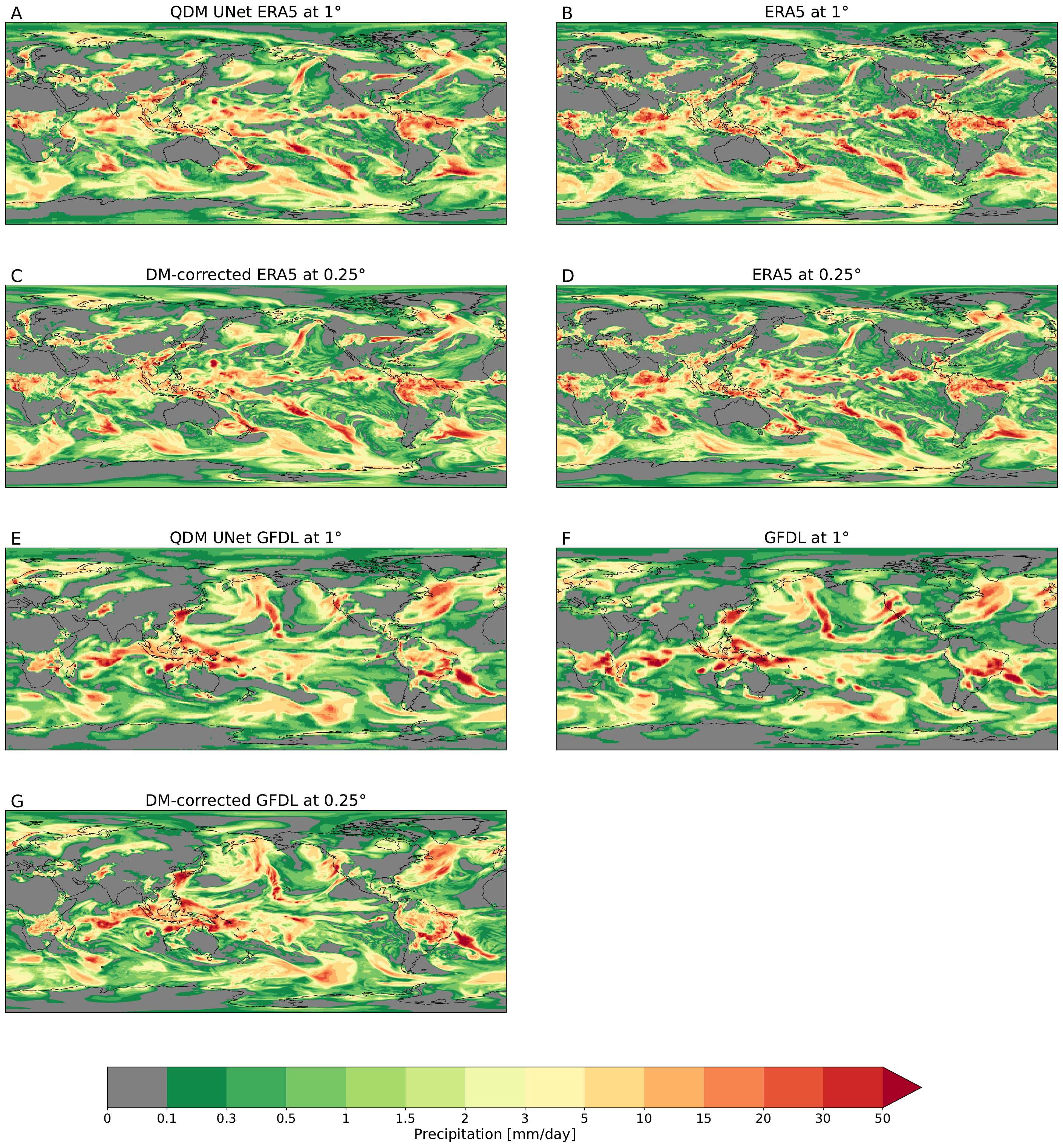}
    \caption{ \textbf{Visual evaluation of our framework's conditional generation capabilities for a representative snapshot.} 
     \textbf{(A–D) Validation against ERA5 data (validation dataset):} (A) Quantile-mapped UNet regression output (1°) conditioned on ERA5 atmospheric variables, (B) Coarse ERA5 precipitation (1°), (C) Diffusion Model generated high-resolution field (0.25°) and (D) ERA5 ground truth (0.25°). Note that in this case, the QDM correction was fitted using the UNet output derived from coarse ERA5 inputs. \textbf{(E–G) Inference with GFDL-ESM:} (E) UNet prediction (1°) conditioned on GFDL atmospheric variables, (F) GFDL precipitation (1°), (G) DM-corrected high-resolution output. 
     The structural consistency of large-scale events across fields A–D and E–G demonstrates that the generated fields are conditioned on the atmospheric state, rather than being unconditional stochastic realizations.
    }
    \phantomsection\label{fig:individual_samples}
\end{figure}

\newpage
\renewcommand{\thefigure}{S\arabic{figure}} 
\section*{Fig. S2}

\begin{figure}[!htb]
    \centering
    \includegraphics[width=\textwidth]{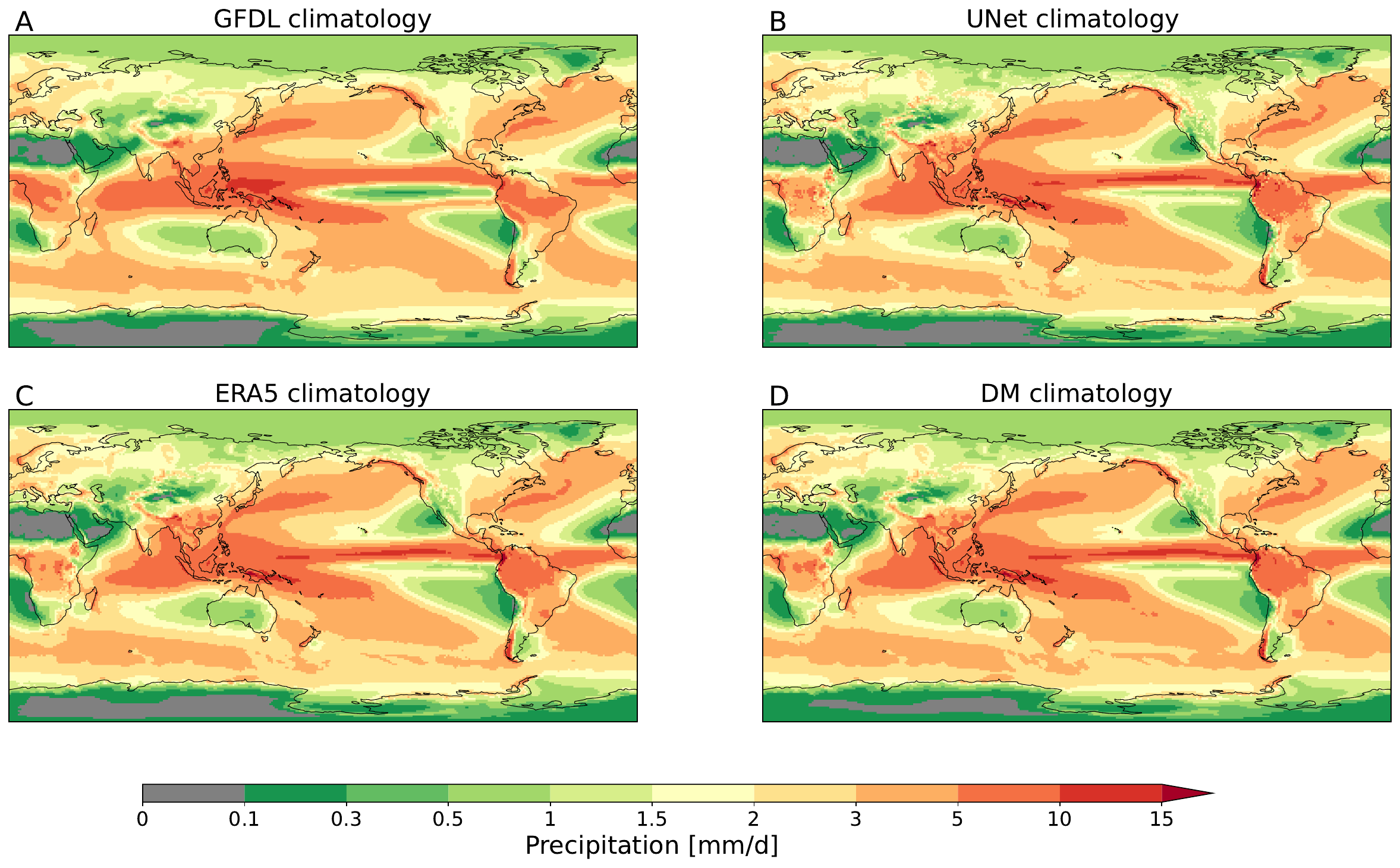}
    \caption{ \textbf{Comparing long term climatologies}. Both our regression model and our diffusion model show climatologies that are closely aligned with ERA5. In contrast, GFDL exhibits a notable ITCZ shift compared to our regression and diffusion models. Note that we interpolated ERA5 and the diffusion model to 1° using average pooling for comparability.}
    \phantomsection\label{fig:climatology}
\end{figure}

\newpage
\renewcommand{\thefigure}{S\arabic{figure}} 
\section*{Fig. S3}

\begin{figure}[!htb]
    \centering
    \includegraphics[width=0.6 \textwidth]{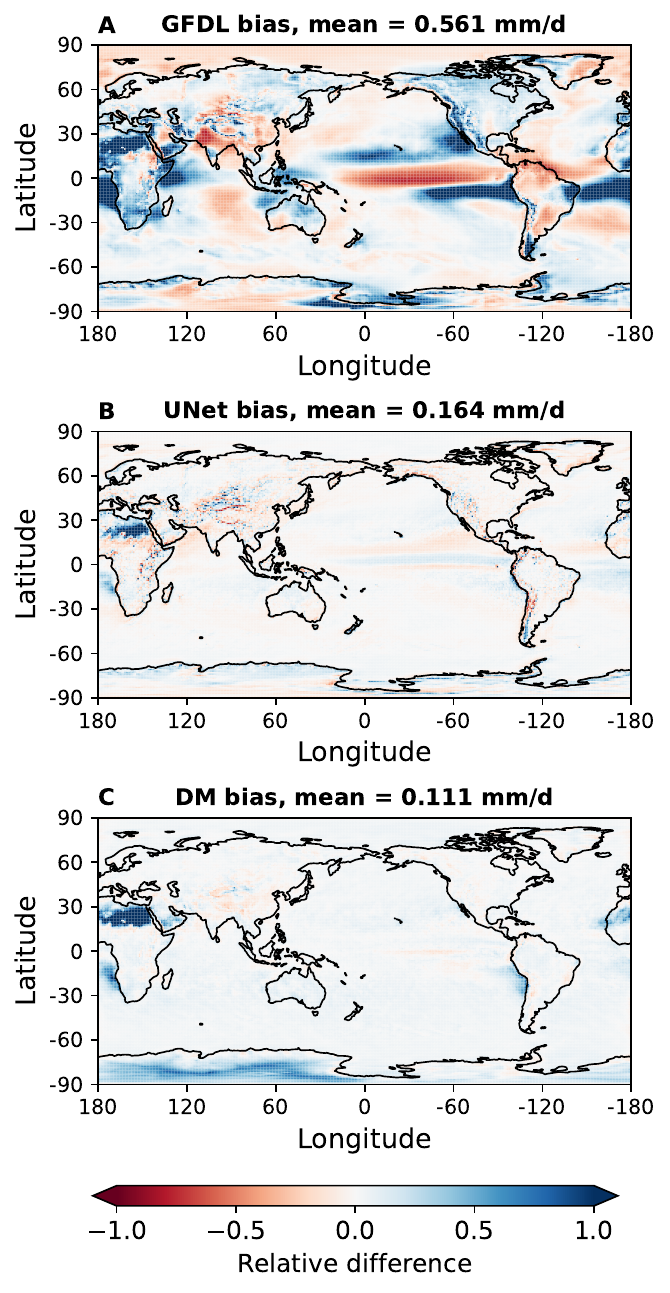}
    \caption{ \textbf{Relative model biases of GFDL, UNet and DM}. The maps show the relative precipitation biases of \textbf{(A)} GFDL, \textbf{(B)} the UNet regression model, and \textbf{(C)} the DM. Positive values (blue) indicate a relative overestimation (wet bias), while negative values (red) indicate a relative underestimation (dry bias). Note that we interpolated ERA5 and the DM to 1° by applying average pooling for fair comparison to GFDL.}
    \phantomsection\label{fig:relative_bias_gfdl}
\end{figure}

\newpage
\renewcommand{\thefigure}{S\arabic{figure}} 
\section*{Fig. S4}

\begin{figure}[!htb]
    \centering
    \includegraphics[width=\textwidth]{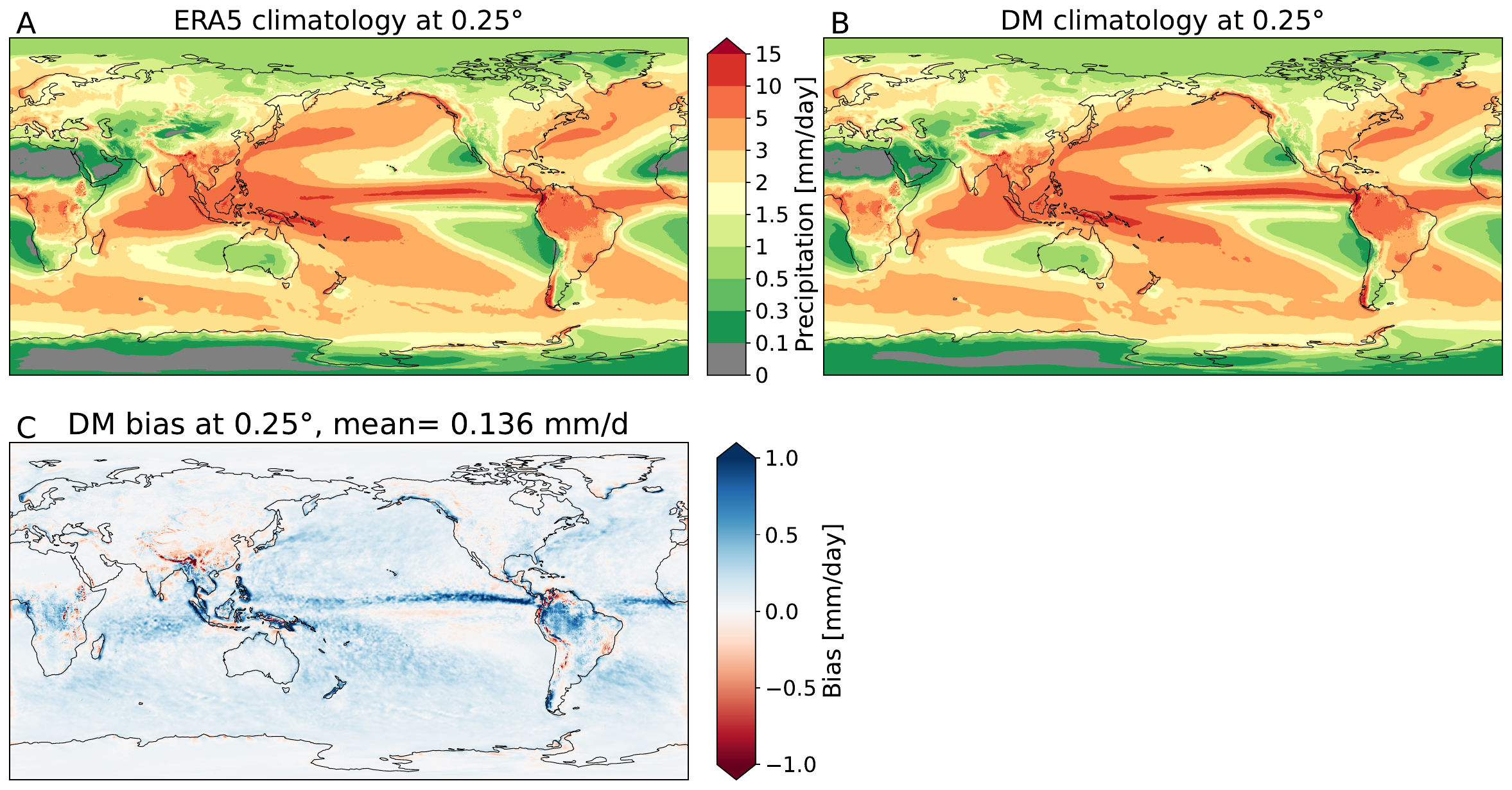}
    \caption{ \textbf{Comparison of the long-term precipitation climatology of ERA5 and the precipitation predicted by our diffusion model at 0.25°}. The climatologies align well, with no indication of a double ITCZ bias in our model. The difference map, along with a low mean absolute bias of 0.136 mm/day, highlights a very minimal bias in our diffusion model.}
    \phantomsection\label{fig:climatology_hr}
\end{figure}

\newpage
\renewcommand{\thefigure}{S\arabic{figure}} 
\section*{Fig. S5}

\begin{figure}[!htb]
    \centering
    \includegraphics[width=\textwidth]{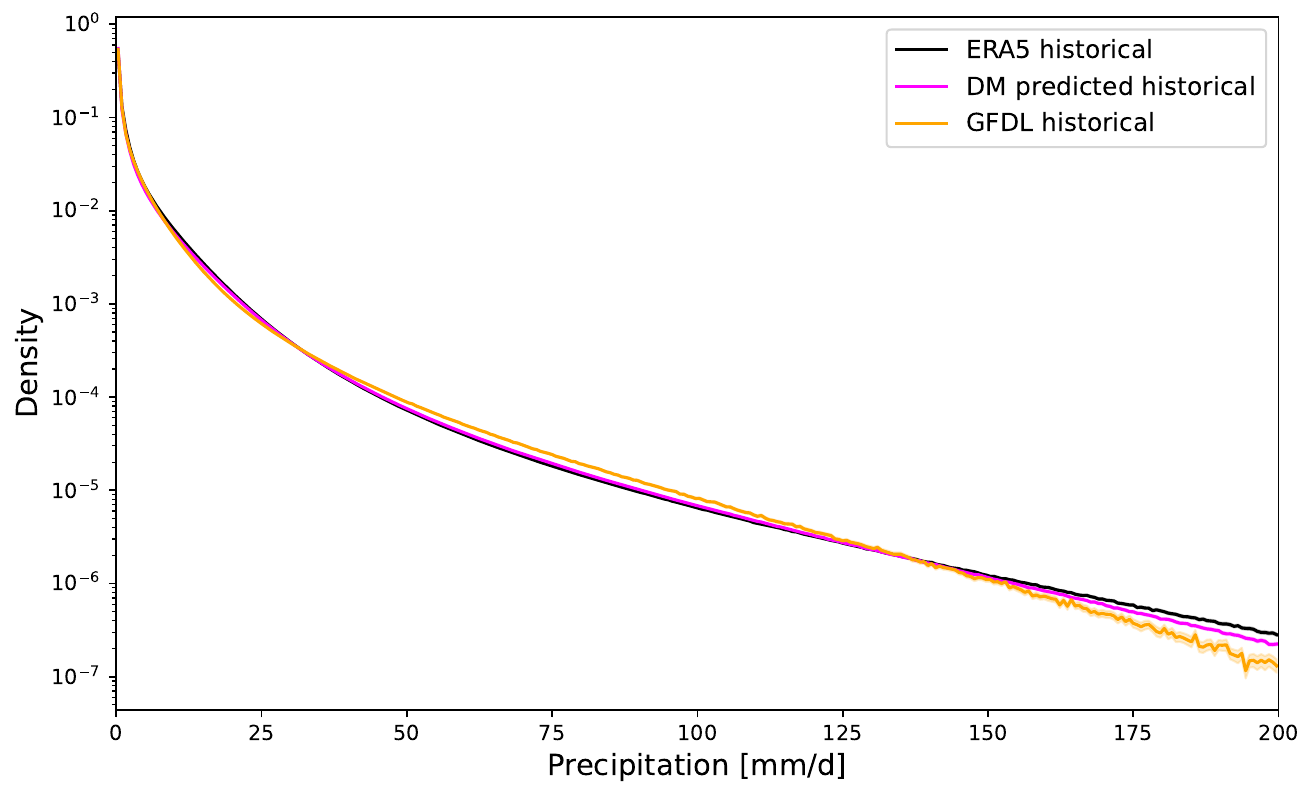}
    \caption{ \textbf{Precipitation histogram with confidence intervals.} The plot compares the daily precipitation distributions of 0.25° ERA5 historical (black), native 1° GFDL historical (yellow), and 0.25° DM predicted historical (magenta) datasets over the historical period. Shaded regions indicate 95$\%$ confidence intervals estimated using a bootstrap analysis with 1000 iterations. }
    \phantomsection\label{fig:bootstrapped_hist}
\end{figure}

\newpage
\renewcommand{\thefigure}{S\arabic{figure}} 
\section*{Fig. S6}

\begin{figure}[!htb]
    \centering
    \includegraphics[width=1\textwidth]{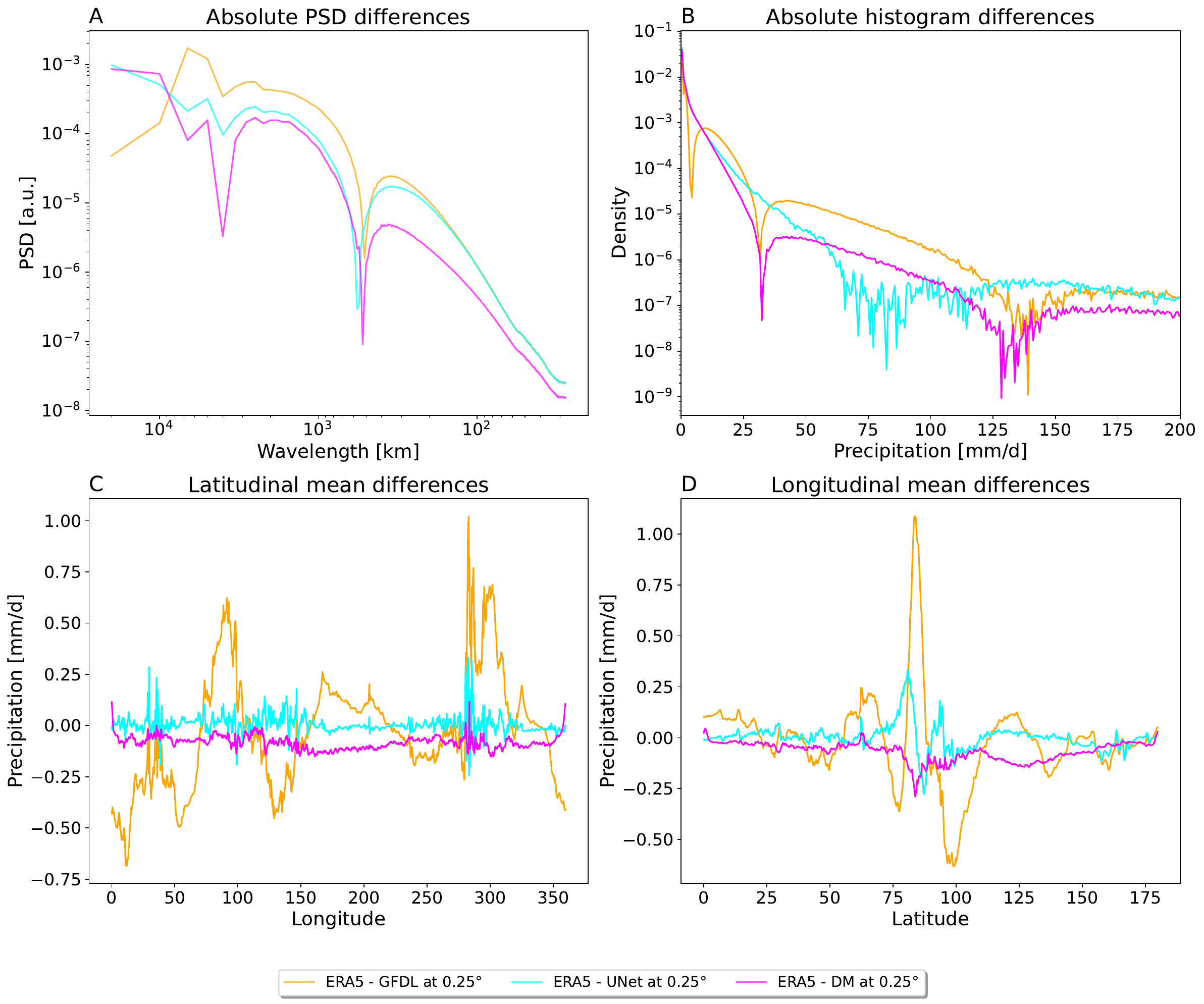}
    \caption{ \textbf{Visualizing the differences between models with respect to ERA5 statistics at 0.25° resolution over 40 years.} We compare the differences of GFDL (bi-linearly interpolated to 0.25°) in yellow, our UNet regression model (bi-linearly interpolated to 0.25°) in cyan, and our diffusion model (DM) at 0.25° in magenta.
    \textbf{(A)} Absolute difference in mean spatial power spectral density (PSD). The diffusion model has the overall smallest difference compared to ERA5. GFDL has much larger differences even in the large spatial scales.
    \textbf{(B)} Absolute differences in the histograms indicating differences in precipitation frequencies. The histogram also shows that our DM has the smallest deviation from ERA5 and is especially superior in the range of 10 mm/day - 60 mm/day. 
    \textbf{(C)} Differences in the longitude profile, given by the data averaged over all longitudes, weighted by the cosine of latitude to account for the varying grid cell area. Both our DM and UNet model have much smaller deviations from ERA5 than GFDL.
    \textbf{(D)} Latitude profile, given by the data averaged over all longitudes. Our DM exhibits the smallest deviation from ERA5, while GFDL shows a pronounced ITCZ bias.
    }
    \phantomsection\label{fig:gfdl_eval_diff}
\end{figure}

\newpage
\renewcommand{\thefigure}{S\arabic{figure}} 
\section*{Fig. S7}

\begin{figure}[!htb]
    \centering
    \includegraphics[width=\textwidth]{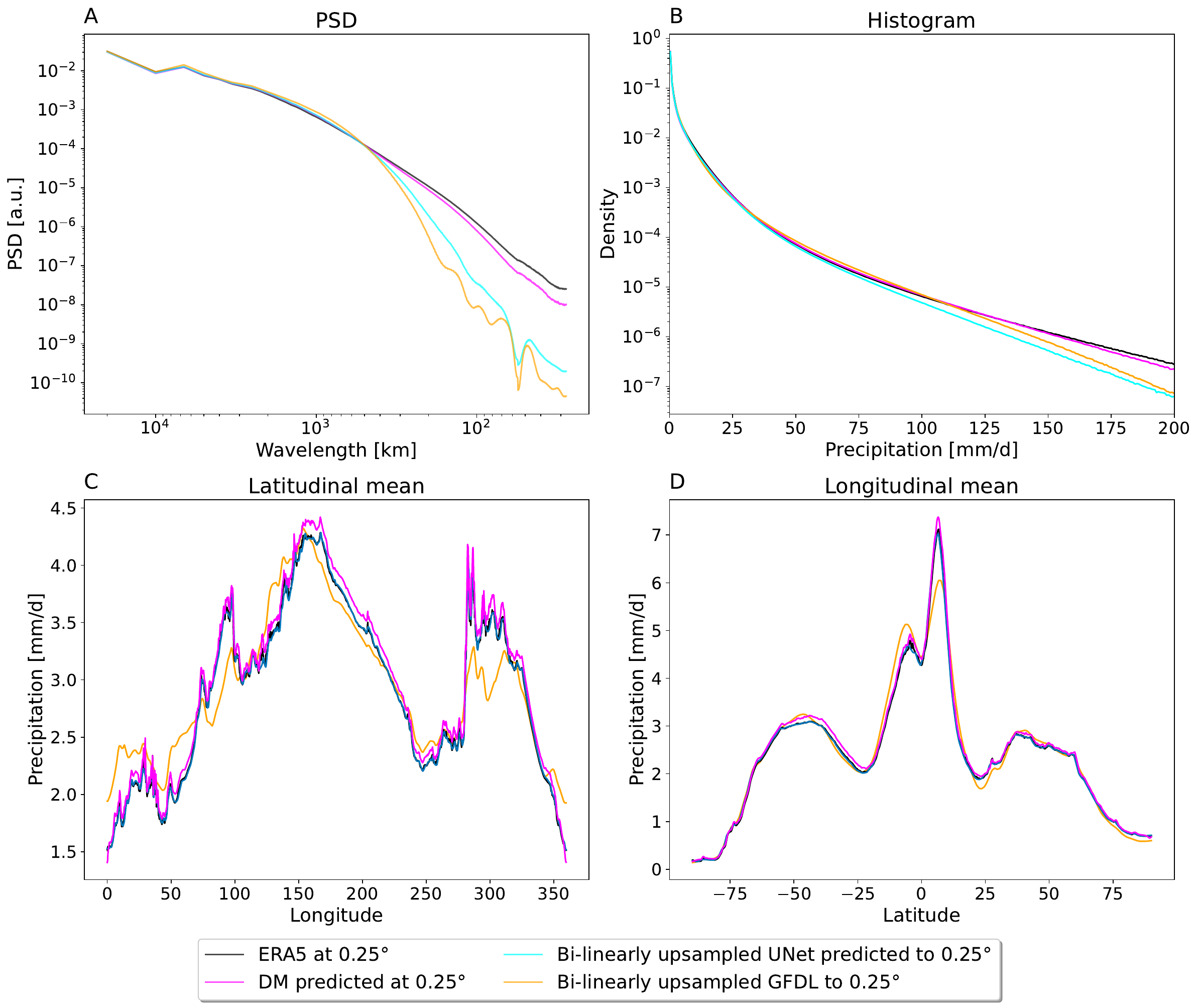}
    \caption{ \textbf{Comparison of the DM against a traditional statistical downscaling baseline.} 
    \textbf{(A)} Mean spatial power spectral density (PSD).
    \textbf{(B)} Histogram indicating the precipitation frequencies.
    \textbf{(C)} Longitude profile, given by the data averaged over all latitudes, weighted by the cosine of latitude to account for the varying grid cell area.
    \textbf{(D)} Latitude profile, given by the data averaged over all longitudes. The statistical baseline (Bi-linearly interpolated QDM) is constructed by quantile delta mapping the UNet regression output followed by bi-linear interpolation to the 0.25° resolution. We compare this to our DM predicted output at 0.25° resolution.
    }
    \phantomsection\label{fig:bilin_bench}
\end{figure}

\newpage
\renewcommand{\thefigure}{S\arabic{figure}} 
\section*{Fig. S8}

\begin{figure}[!htb]
    \centering
    \includegraphics[width=\textwidth]{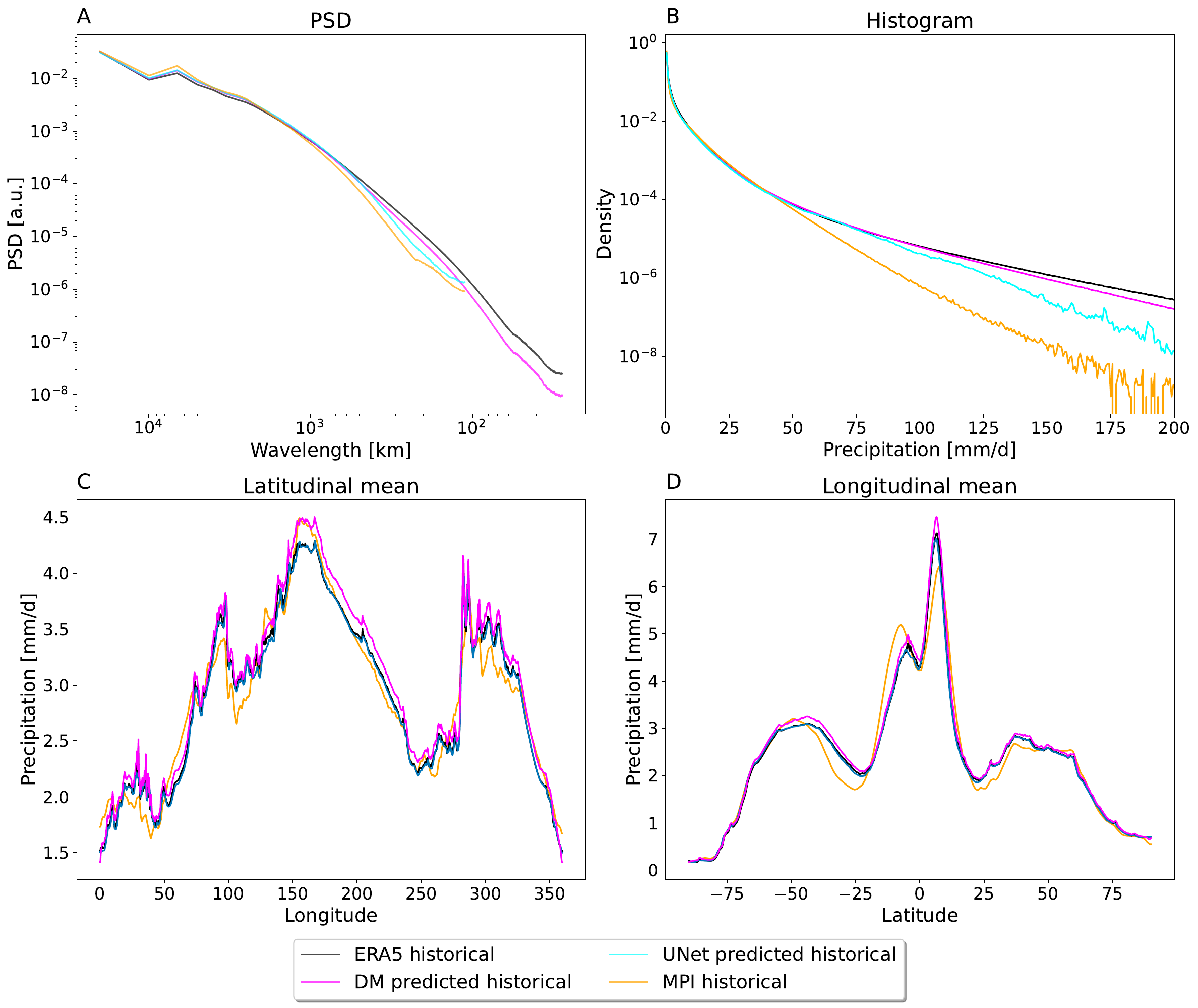}
    \caption{ \textbf{Reproduction of the statistics of ERA5 at 0.25° resolution over 40 years for the MPI-ESM.} 
    \textbf{(A)} Mean spatial power spectral density (PSD).
    \textbf{(B)} Histogram indicating the precipitation frequencies.
    \textbf{(C)} Longitude profile, given by the data averaged over all latitudes, weighted by the cosine of latitude to account for the varying grid cell area.
    \textbf{(D)} Latitude profile, given by the data averaged over all longitudes.   
    For panels (B-C), we bi-linearly interpolate the MPI-ESM and our regression model predictions of precipitation (orange/cyan) from 1° to 0.25° to compare against the native 0.25° ERA5 and DM fields.
    }
    \phantomsection\label{fig:mpi_benchmark}
\end{figure}

\newpage
\renewcommand{\thefigure}{S\arabic{figure}} 
\section*{Fig. S9}

\begin{figure}[!htb]
    \centering
    \includegraphics[width=\textwidth]{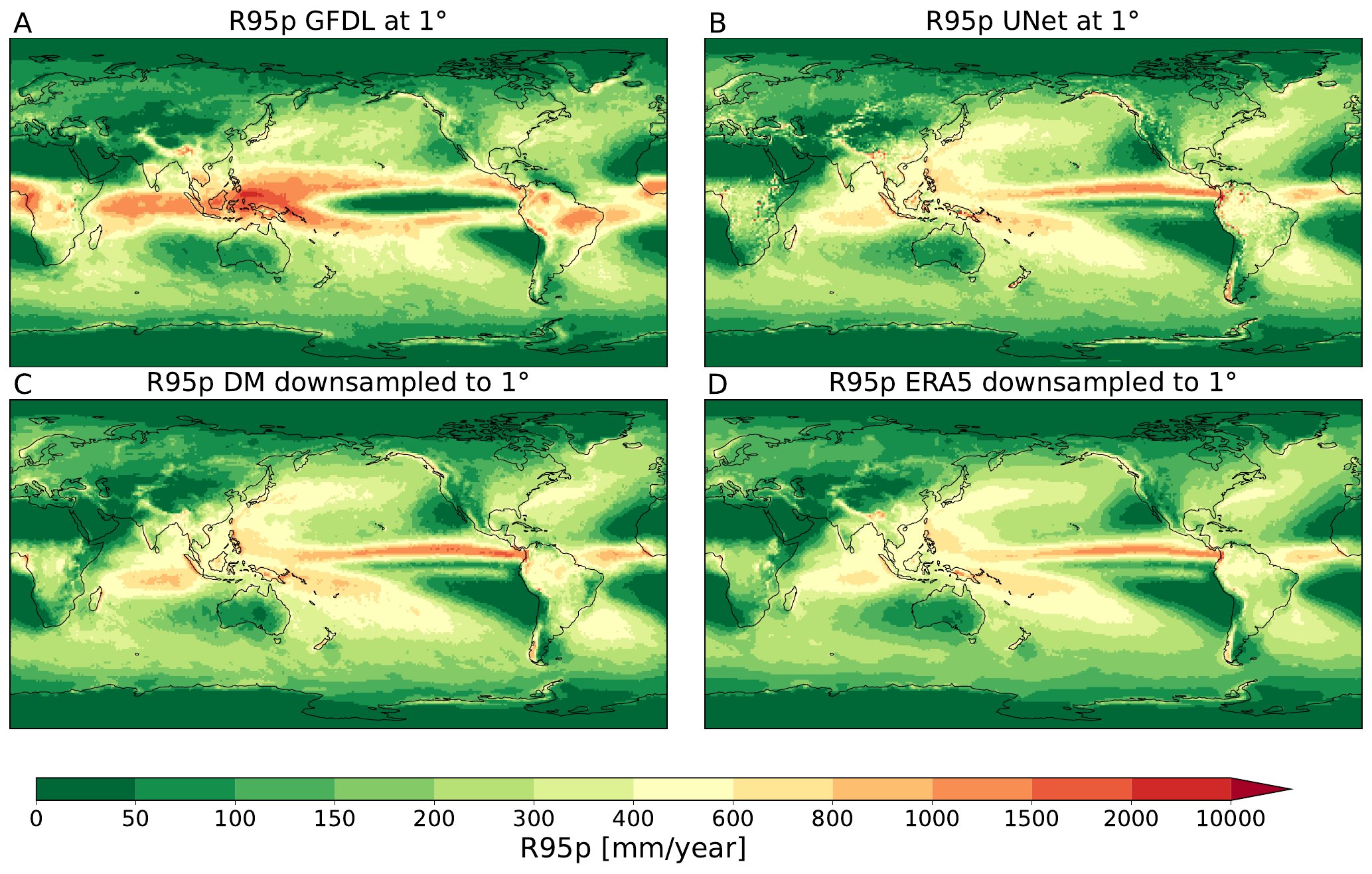}
    \caption{\textbf{Evaluation of extreme event coverage for the historical period.} \textbf{(A-D)}: Climatology of R95p (annual total precipitation from days exceeding the 95th percentile) for the historical period simulated by (A) UNet at 1°, (B) GFDL at 1°, and (C) the diffusion model (DM) interpolated to 1°, compared to (D) ERA5 reanalysis interpolated to 1°. The GFDL model exhibits strong wet biases in the tropics (A), whereas the UNet and diffusion models show reduced biases (C, D), aligning well with the spatial patterns of ERA5.
    }
    \phantomsection\label{fig:r95p_clima}
\end{figure}

\newpage
\renewcommand{\thefigure}{S\arabic{figure}} 
\section*{Fig. S10}

\begin{figure}[!htb]
    \centering
    \includegraphics[width=\textwidth]{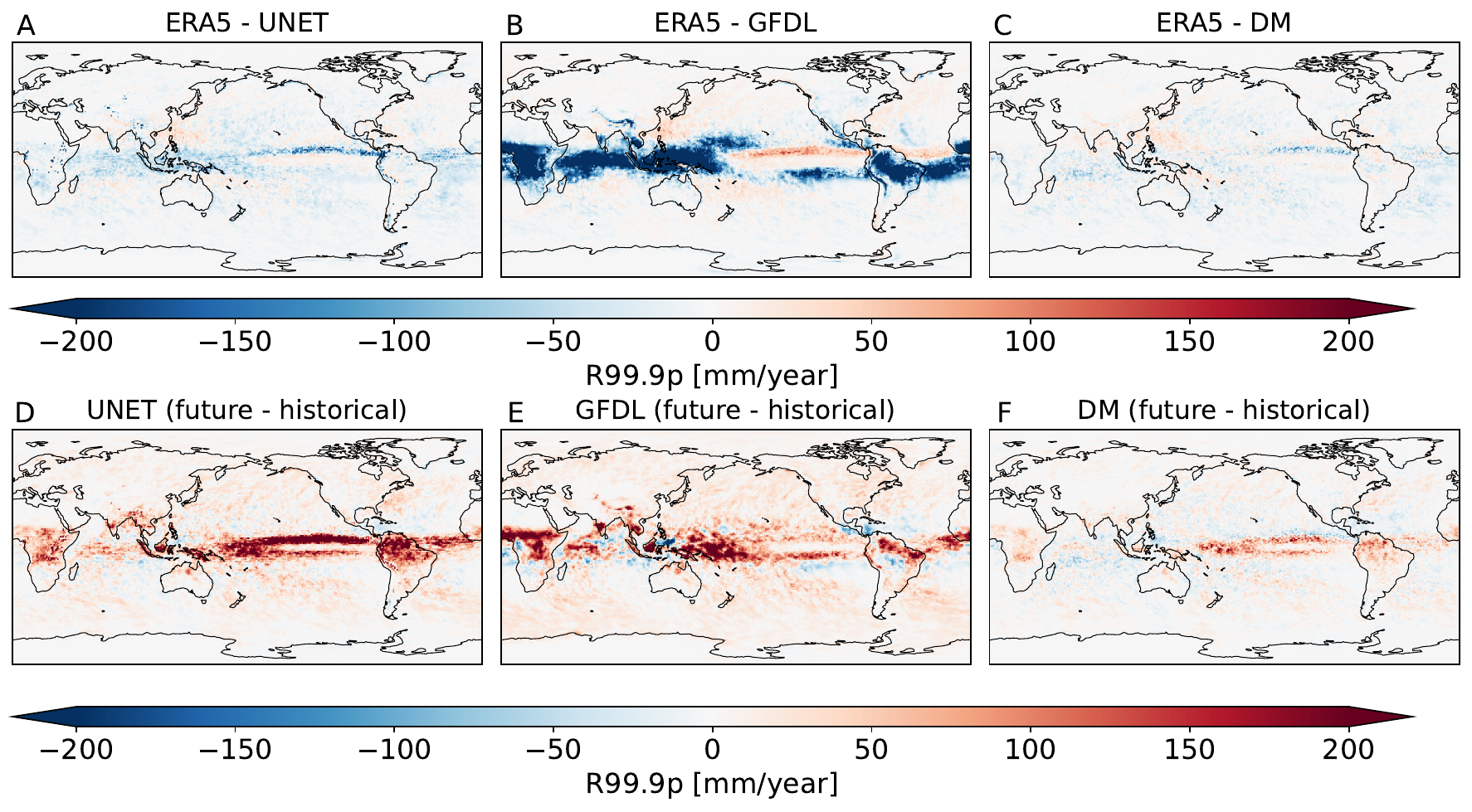}
    \caption{ \textbf{Evaluation of rare extreme event coverage (R99.9p}). Similar to Fig. \ref*{fig:r95p_dif} we evaluate the total precipitation from days exceeding the 99.9th percentile of daily rainfall (R99.9p). ERA5 and the DM were interpolated to 1° to match the GFDL resolution. (A-C) Bias in historical data (1980 – 2005) relative to ERA5. Negative values (blue) indicate overestimation of extreme precipitation by the model. The DM (C) effectively reduces the strong wet bias present in the GFDL model (B), showing a spatial error pattern comparable to the quantile delta mapped UNet (A). (D-F) Projected changes in R99p between the late 21st century (2075 – 2100, SSP3-7.0) and the historical period. All models predict that these rarer extremes in the tropics will become more pronounced.}
    \phantomsection\label{fig:r99_difference}
\end{figure}

\newpage
\renewcommand{\thefigure}{S\arabic{figure}} 
\section*{Fig. S11}

\begin{figure}[!htb]
    \centering
    \includegraphics[width=\textwidth]{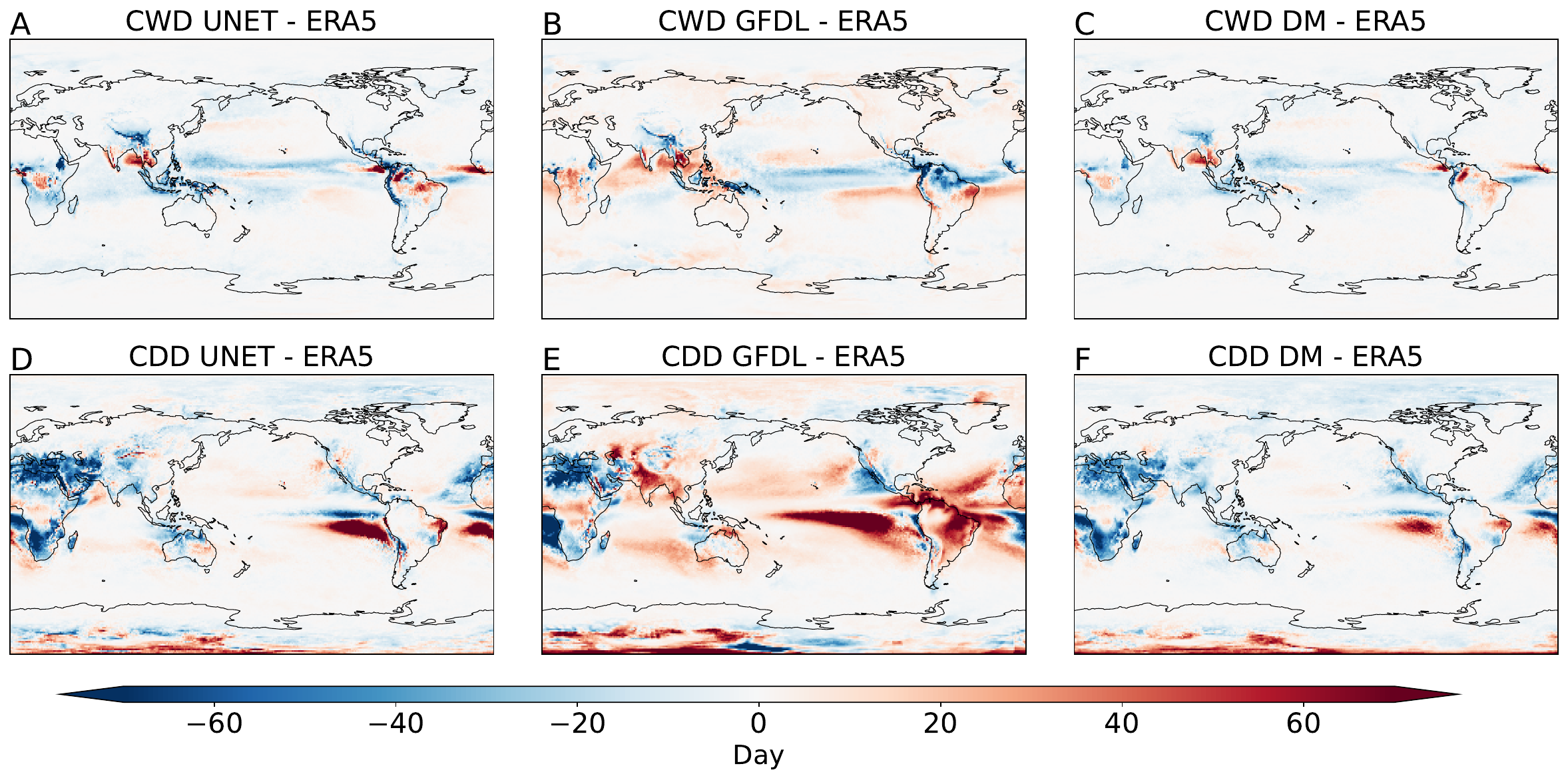}
    \caption{ \textbf{Comparing consecutive wet days (CWD) and consecutive dry days for the historical period 1980-2005.} We show the CWD-differences (top row) and CDD-differences (bottom row) compared to ERA5. All comparisons were performed at 1° resolution. Panels (A–C) show CWD anomalies from the regression model, GFDL, and the diffusion model, respectively, while panels (D–F) display CDD anomalies for the same three datasets. Red areas indicate an increase relative to ERA5, and blue areas a decrease. Our DM has overall the smallest bias compared to ERA5, especially improving the wet-bias around the tropics of GFDL}
    \phantomsection\label{fig:cdd_cwd_dif}
\end{figure}

\newpage
\renewcommand{\thefigure}{S\arabic{figure}} 
\section*{Fig. S12}

\begin{figure}[!htb]
    \centering
    \includegraphics[width=\textwidth]{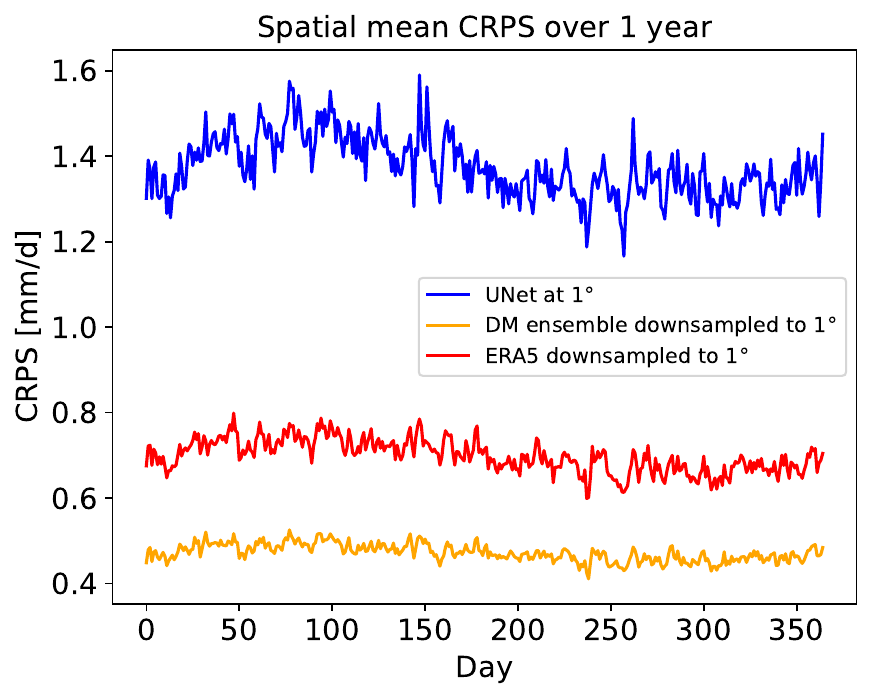}
    \caption{ \textbf{ Evaluation of spatial mean CRPS at the coarse 1° scale}. Time series of the spatially averaged Continuous Ranked Probability Score (CRPS) over one test year, evaluated at 1° resolution. To obtain these scores, the predictions from the 0.25° DM ensemble and the ERA5 data, were interpolated to 1°. The quantile delta mapped UNet regression output is at native 1° resolution. QDM was fitted using the UNet output derived from coarse ERA5 inputs. The absolute CRPS values decrease for all models compared to the 0.25° scale, due to the reduction of variance through the upscaling. The relative performance is the same as in Fig. \ref*{fig:crps}, with the DM having the lowest (best) score.}
    \phantomsection\label{fig:crps_1degre}
\end{figure}

\newpage
\renewcommand{\thefigure}{S\arabic{figure}} 
\section*{Fig. S13}

\begin{figure}[!htb]
    \centering
    \includegraphics[width=\textwidth]{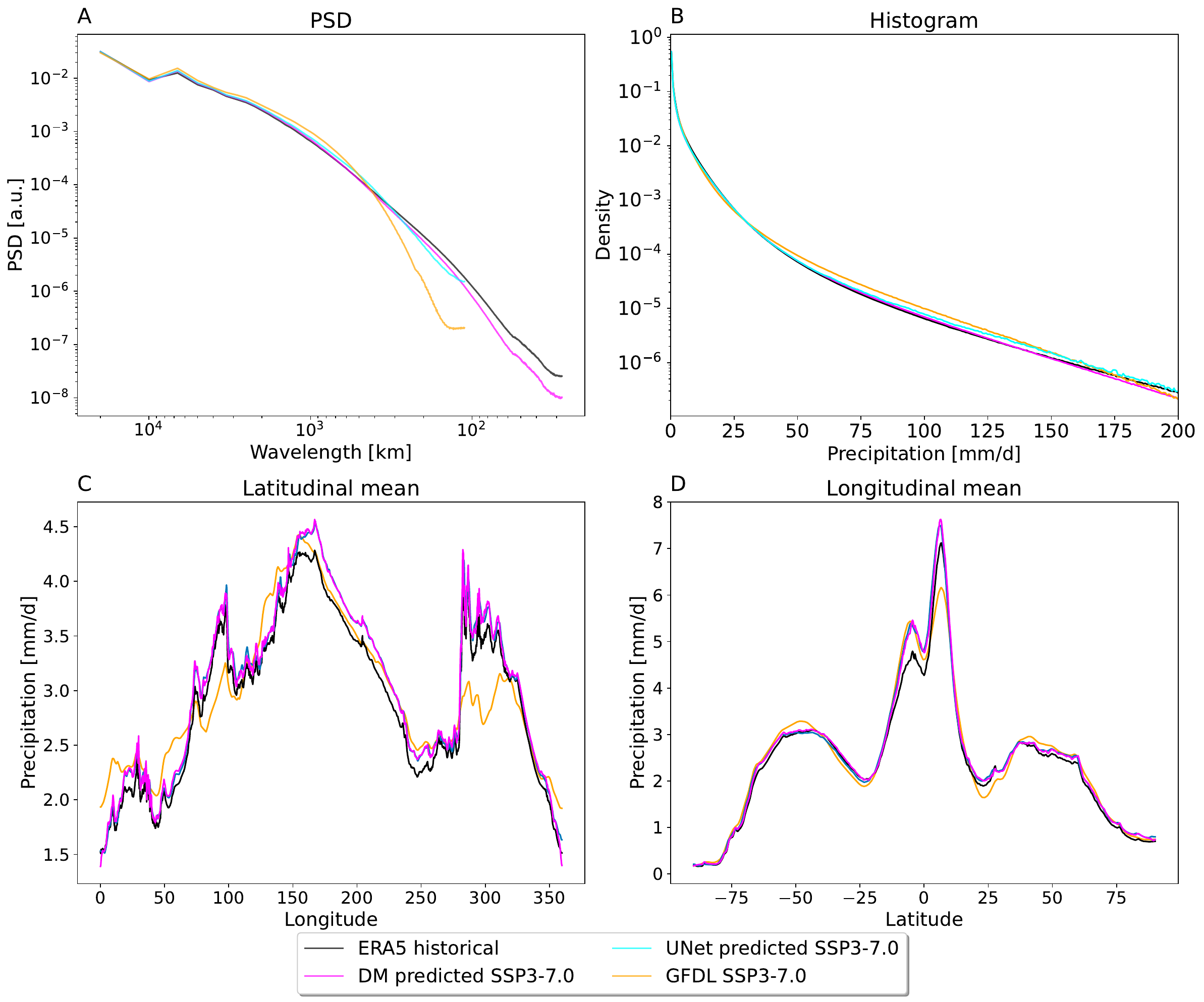}
    \caption{\textbf{Comparing the statistics of different models in the SSP3-7.0 scenario.} 
    The mean spatial power spectral density (PSD) shows that our diffusion model (magenta) improves the small-scale spatial variability and is able to increase the resolution of our regression model (cyan) from 1° to 0.25° \textbf{(A)}. There is a large agreement in the large spatial scales and only a slight disagreement compared to ERA5 (black) for the very small scales. GFDL (yellow) and our regression model (both at 1°) have mismatches throughout the spectrum the get increasingly larger for smaller-scales in the case of GFDL. 
    In the following comparisons, we bi-linearly interpolated the GFDL and our regression model’s predicted precipitation (orange/cyan) from 1° to the 0.25° resolution of ERA5 and the diffusion model.
    The histogram \textbf{(B)} shows large improvements of our models compared to GFDL with slight deviations from the HR ERA5 reference data only for the extreme precipitation events. The regression model is always further away from ERA5 than the DM, expect for the very extreme precipitation.
    The Latitude/Longitude \textbf{(C)}/\textbf{(D)} profiles are given by the averaged longitudes/latitudes show that our regression model and our diffusion model approximate the latitude and longitude profile of the original ERA5 reference data much better than GFDL. The longitude profile is weighted by the cosine of latitude to account for the varying grid cell area.}
    \phantomsection\label{fig:ssp_statisitcs}
\end{figure}

\newpage
\renewcommand{\thefigure}{S\arabic{figure}} 
\section*{Fig. S14}

\begin{figure}[!htb]
    \centering
    \includegraphics[width=\textwidth]{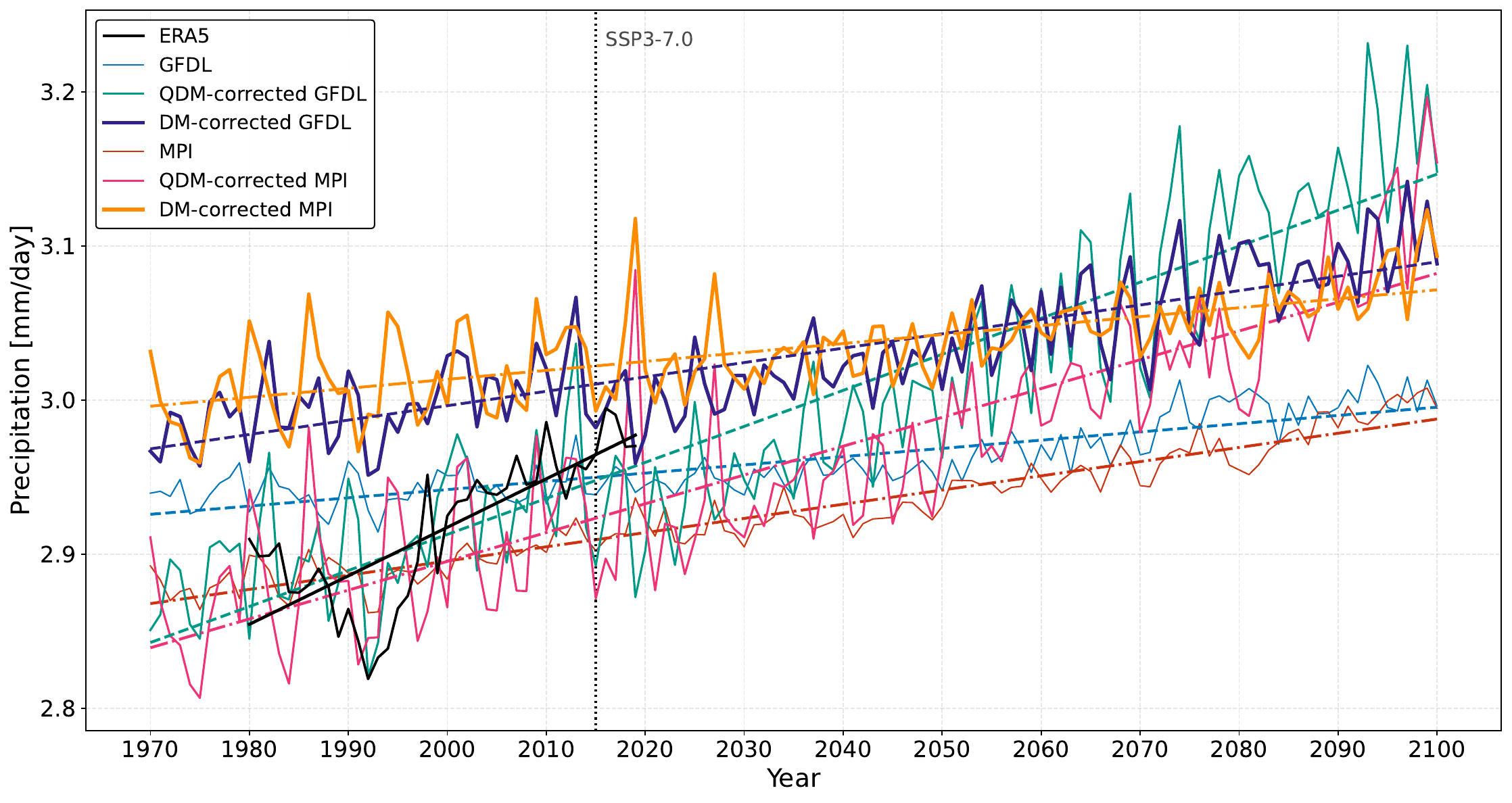}
    \caption{ \textbf{Precipitation trend preservation.} Time series of the annual precipitation for the period 1970–2100, spanning both historical and SSP3-7.0 scenarios. The plot compares the raw ESM output with the quantile delta mapping-corrected UNet outputs and our diffusion model. For reference the historical period shows ERA5 data. The dashed lines represent the linear regression trends for each model, and the vertical dotted line marks the transition to the future scenario. While QDM correction exhibits high inter-annual variance, the DM reduces this. The DM-corrected outputs preserve the climatological trends of their respective driving ESMs across the entire period, applying a mean correction without destroying the underlying climate change signal.
    }
    \phantomsection\label{fig:full_trend}
\end{figure}

\newpage
\renewcommand{\thefigure}{S\arabic{figure}} 
\section*{Fig. S15}

\begin{figure}[!htb]
    \centering
    \includegraphics[width=0.6\textwidth]{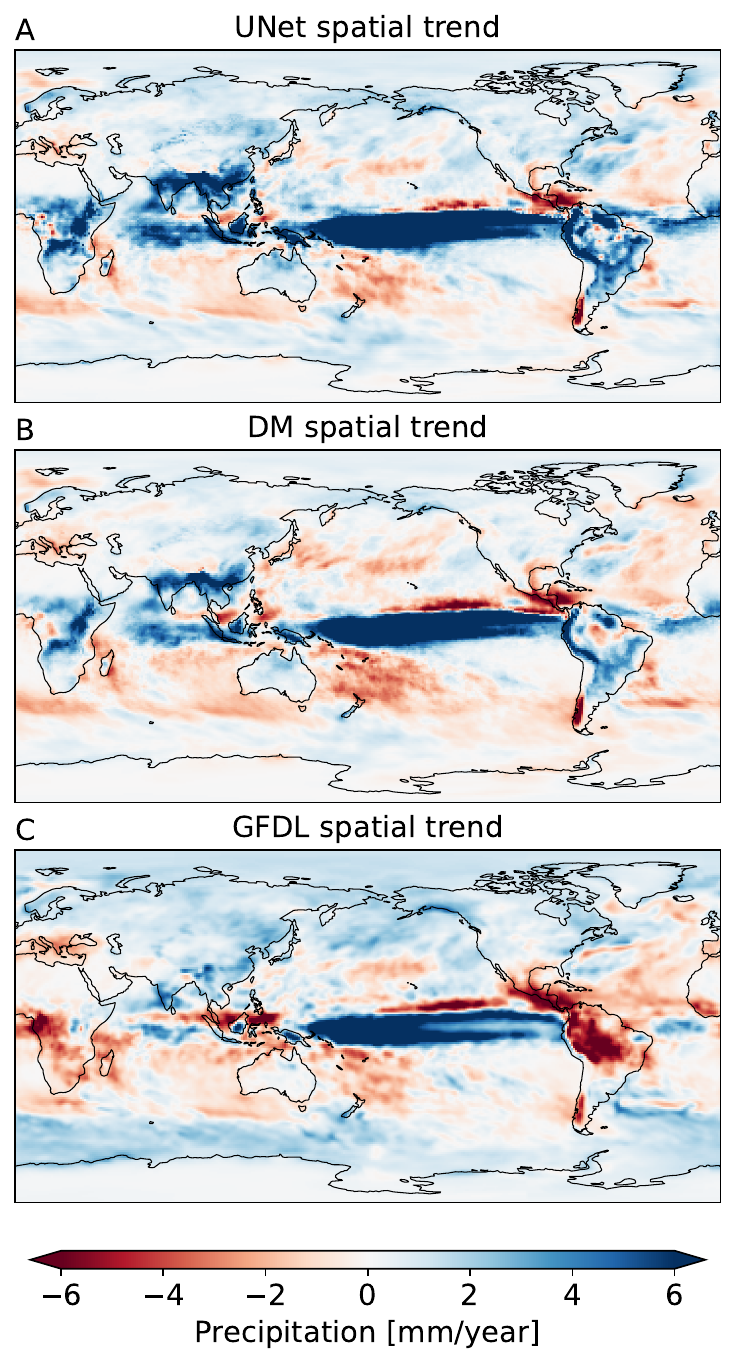}
    \caption{ \textbf{Our diffusion model (DM) preserves the spatial features of the trend from the UNet under the SSP3-7.0 scenario}. \textbf{(A-C)} show the trend at each location between 2015 and 2100 for the UNet, DM, and GFDL as a reference. For comparison, we upscaled the DM precipitation fields to the 1° resolution of the UNet and GFDL. While the DM preserves the spatial trend patterns of the UNet, it adjusts the overall trend to be smaller.
    }
    \phantomsection\label{fig:spatial_trend_ssp370}
\end{figure}

\newpage
\renewcommand{\thefigure}{S\arabic{figure}} 
\section*{Fig. S16}

\begin{figure}[!htb]
    \centering
    \includegraphics[width=0.6 \textwidth]{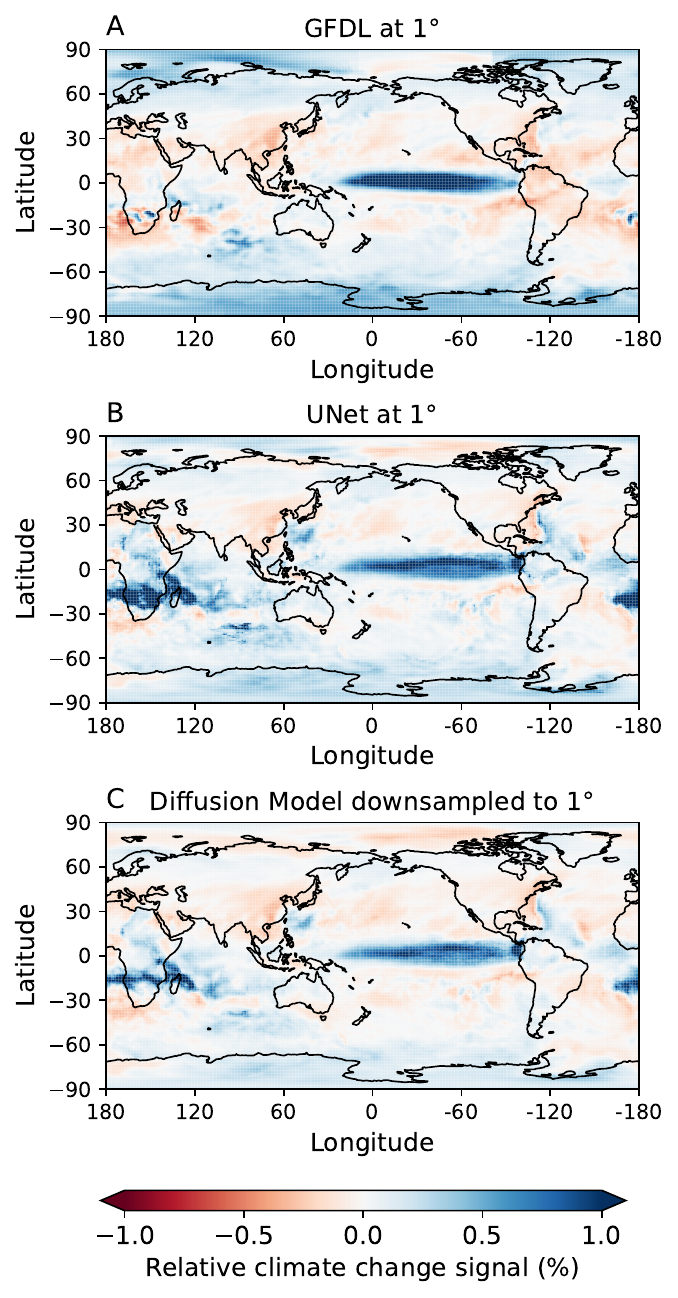}
    \caption{ \textbf{Relative precipitation changes (\%)}. All changes compare the late 21st century (2075-2100) under the SSP3.7.0 scenario and the present‐day period (1980–2005). \textbf{(A)} GFDL precipitation, \textbf{(B)} the UNet regression model, and \textbf{(C)} our diffusion model (interpolated to 1°). The sign of the change is mostly consistent across the maps. The diffusion model predicts a slightly weaker change than the regression model but maintains similar spatial patterns.
    }
    \phantomsection\label{fig:rel_dif_past_future}
\end{figure}

\end{document}